\documentclass[mnsc,nonblindrev]{informs3aa} 

\usepackage{natbib}
 \bibpunct[, ]{(}{)}{,}{a}{}{,}%
 \def\bibfont{\small}%
\TheoremsNumberedThrough     
\ECRepeatTheorems

\EquationsNumberedThrough    

\MANUSCRIPTNO{MS-0001-1922.65}

\usepackage{amsmath}

\usepackage{amssymb}
\usepackage{algorithm}
\usepackage[noend]{algpseudocode}
\usepackage{mathtools}

\usepackage{marvosym}
\usepackage{dsfont}
\usepackage[nopar]{lipsum}
\usepackage{enumitem}
\usepackage{graphicx}
\usepackage{caption}
\usepackage{subcaption}
\usepackage{multirow}
\usepackage{soul}
\usepackage{bbm}
\usepackage{booktabs}
\def\undertilde#1{\mathord{\vtop{\ialign{##\crcr
$\hfil\displaystyle{#1}\hfil$\crcr\noalign{\kern1.5pt\nointerlineskip}
$\hfil\tilde{}\hfil$\crcr\noalign{\kern1.5pt}}}}}
\usepackage{tabularray}
\usepackage[framemethod=tikz]{mdframed}
\usepackage{placeins}
\usepackage{tablefootnote}
\usepackage{natbib}
\usepackage{url}

\usepackage{hyperref}
\hypersetup{colorlinks=true,urlcolor=blue, citecolor=blue,linkcolor=blue,bookmarksopen=false,draft=false}

\def\QED{\hfill \quad{\bf Q.E.D.}\medskip}

\begin{document}

\RUNAUTHOR{Zhalechian, Saghafian, and Robles}
\RUNTITLE{Enhancing the FDA’s Medical Device Clearance}
\TITLE{\Large Harmonizing Safety and Speed: A Human-Algorithm Approach to Enhance the FDA’s Medical Device Clearance Policy}

\ARTICLEAUTHORS{%
\AUTHOR{Mohammad Zhalechian$^a$, Soroush Saghafian$^b$, Omar Robles$^c$}
\AFF{$^a$Kelley School of Business, Indiana University, Bloomington, IN, \EMAIL{ mzhale@iu.edu}}
\AFF{$^b$Harvard Kennedy School, Harvard University, Cambridge, MA, \EMAIL{soroush\_saghafian@hks.harvard.edu}}
\AFF{$^c$Emerging Health Consulting, Armonk, NY, \EMAIL{ orobles@emerginghealthllc.com}}
} 

\ABSTRACT{The United States Food and Drug Administration's (FDA's) 510(k) pathway allows manufacturers to gain medical device approval by demonstrating substantial equivalence to a legally marketed device. However, the inherent ambiguity of this regulatory procedure has been associated with high recall among many devices cleared through this pathway, raising significant safety concerns. In this paper, we develop a combined human-algorithm approach to assist the FDA in improving its 510(k) medical device clearance process by reducing recall risk and regulatory workload. We first develop machine learning methods to estimate the risk of recall of 510(k) medical devices based on the information available at the time of submission. We then propose a data-driven clearance policy that recommends acceptance, rejection, or deferral to FDA's committees for in-depth evaluation. We conduct an empirical study using a unique dataset of over 31,000 submissions that we assembled based on data sources from the FDA and Centers for Medicare and Medicaid Service (CMS). \textcolor{black}{Compared to the FDA's current practice, which has a recall rate of 10.3\% and a normalized workload measure of 100\%, a conservative evaluation of our policy shows a 32.9\% improvement in the recall rate and a 40.5\% reduction in the workload. Our analyses further suggest annual cost savings of approximately \$1.7 billion for the healthcare system driven by avoided replacement costs, which is equivalent to 1.1\% of the entire United States annual medical device expenditure. Our findings highlight the value of a holistic and data-driven approach to improve the FDA's current 510(k) pathway.}}
\HISTORY{Accepted by \textit{Management Science}.\footnotemark}

\maketitle

{\renewcommand{\thefootnote}{*}%
\footnotetext{The published article is available at
\url{https://pubsonline.informs.org/doi/10.1287/mnsc.2024.06477}.}}
\setcounter{footnote}{0}

\section{Introduction}
The medical device approval pathway is a function of the level of control necessary to provide reasonable assurance of a device’s safety and effectiveness (\citealt{FDApathway}). In general, devices posing a greater degree of risk are denoted by class and face greater regulatory controls. Although there are four common types of approval pathways (\citealt{FDAmarket}), the vast majority of medical devices are cleared under the \textit{Premarket Notification 510(k) pathway}. For the year 2022, over 3,000 submissions were cleared under the 510(k) pathway (\citealt{510(k)pathway}), while fewer than two hundred devices were cleared under the Humanitarian Device Exemption, Premarket Approval, and De Novo pathways combined (\citealt{HDE}, \citealt{2022devices}, and \citealt{DeNovo}).

The 510(k) pathway was developed both to reduce the burden for device manufacturers bringing medium-to-low risk (Class II and I) devices to market and to address the limited resources of the United States Food and Drug Administration (FDA) (\citealt{kramer2023quantitative}). Devices cleared under the 510(k) pathway must demonstrate that the device is “substantially equivalent” to a legally marketed device, commonly referred to as a \textit{predicate device} (\citealt{premarketnotif}). A predicate device can be (a) any device that is legally marketed prior to May 28, 1976 (a preamendments device), for which clinical testing was not required, (b) any device that was reclassified from Class III (high risk) to Class II or Class I (medium-to-low-risk), or (c) any device that was also found to be substantially equivalent under the 510(k) pathway (\citealt{FDApathway}). \textcolor{black}{An applicant seeking FDA clearance for a device through the 510(k) pathway is required to select at least one predicate but may choose multiple predicates at their discretion. The FDA provides guidance on how to find and effectively use predicate devices. In general, manufacturers are asked to identify a predicate that is most similar to the device under review with respect to indications for use and technological characteristics (\citealt{FDApredicateselection}). However, manufacturers may choose to reference multiple predicates when combining features from different devices with the same intended use, seeking clearance for a device with multiple intended uses, or requesting multiple indications for use under a shared intended use (\citealt{FDAguidance}). The applicant is not required to identify the universe of predicate devices that meet this standard.} Substantial equivalence, in turn, occurs when the device has the same intended use and technological characteristics, or has different technological characteristics but is as safe and effective as the predicate. The FDA determines whether the device is as safe and effective as the predicate device by reviewing the scientific methods used to evaluate differences in technological characteristics and performance data (\citealt{guidance}).

The 510(k) pathway remains the principal route through which most moderate-to-low risk devices reach the U.S. market, offering patients timely access to incremental innovation. Nonetheless, the pathway has attracted considerable scrutiny. For example, the National Academy of Medicine stated that “the 510(k) process cannot be transformed into a premarket evaluation of safety and effectiveness as long as the standard for clearance is substantial equivalence to any previously cleared device” (\citealt{premarketnotif}), and the U.S. Supreme Court similarly concluded that “the 510(k) process is focused on equivalence, not safety” (\citealt{US_Supreme_Court}). More recently, the FDA itself has acknowledged these limitations, emphasizing the need for alternative approaches to demonstrate that a device is as safe and effective as a legally marketed device (\citealt{challoner2011medical}).
These concerns are reflected in post-market outcomes, specifically device recalls. The FDA classifies an action as a recall when the manufacturer corrects or removes a device that is out of compliance with FDA regulations, typically due to faults or health hazards (\citealt{FDARecall}). 
Moreover, these failures are not always isolated incidents; a single device can accumulate multiple recall events over its commercial life. Several studies have also highlighted the potential drawbacks associated with the 510(k) pathway, indicating that 71\% of devices recalled by the FDA for safety concerns between 2005 and 2009 had initially been cleared through the 510(k) pathway (\citealt{zuckerman2011medical}), or that 11\% of devices cleared via the 510(k) pathway between 2003 and 2018 had later been subject to Class I or II recalls (\citealt{everhart2023association}). Despite these concerns, there is currently a lack of evidence-based understanding of how the FDA can improve its 510(k) pathway. This is evidenced by the FDA's 2023 solicitation for feedback  (\citealt{FDABestPractices} and \citealt{FDAAILML}) and its 2018 Medical Device Safety Action Plan for modernizing the 510(k) pathway (\citealt{FDASafetyActionPlan}). 

\color{black}
Our main goal in this paper is to develop a data-driven approach that can assist the FDA in improving its 510(k) pathway. To this end, we note that developing a highly accurate predictive model of recall risk can enable the FDA to pinpoint devices that might require more rigorous examination or necessitate stronger evidential support before approval, ultimately helping the FDA to mitigate unnecessary harm to patients. Developing such a prediction model requires access to large data containing comprehensive information on both applicant devices and predicates as well as detailed data engineering processes to generate useful insights. This level of detail cannot be directly obtained using publicly available FDA datasets and requires significant additional data collection and pre-processing efforts. 

However, developing a predictive model alone is unlikely to be sufficient for improving the FDA’s 510(k) pathway. The first challenge is related to the nuanced differences of some 510(k) devices, which may make it hard for a policy fully relying on a machine learning (ML) model to be effective. There is a vast literature in support of the interaction between ML models and humans to achieve better overall performance. Thus, we hypothesize that a combined human-algorithm approach to evaluating medical devices can significantly improve FDA's 510(k) pathway. The second challenge is that any such policy must also consider \textcolor{black}{the limited resources} available to the FDA to perform further examination when a device is deemed risky for approval by the model. The FDA receives a high volume of 510(k) submissions each year and has the goal of making a decision within 90 days. The sheer number of submissions can strain the FDA's resources, resulting in a potential backlog and delays in the review process. This backlog may limit the depth and rigor of the review, potentially impacting the thoroughness of the evaluation and the ability to identify potential risks or issues for a variety of devices. Thus, to rigorously improve the 510(k) pathway, one needs to also develop an evaluation policy guideline that not only can benefit from the predicted risk of recall, but can also take into account FDA's limited resources for further evaluation of devices that are not a clear-cut for approval or rejection decisions based on the model's predicted risk. 

\subsection{Overview and Contributions}
We introduce a \textit{data-driven clearance policy} aimed at assisting the FDA in improving its 510(k) pathway. Our proposed policy creates a combined human-algorithm approach by deferring some decisions to human experts and others to a well-trained ML model. Specifically, our policy generates recommendations for direct approval, rejection, or further evaluation by human experts using a variety of the applicant device's characteristics as well as those of its predicates. \textcolor{black}{We note that our proposed policy is designed to support the FDA's substantive review of devices that have passed the initial administrative and eligibility checks (e.g., verification of intended use and technological characteristics). Specifically, our predictive model assesses the safety risk of devices that have established a valid predicate basis.} Our work is based on \textcolor{black}{close collaboration} and extensive discussions with a collaborator (and co-author of this paper) who has been directly involved in numerous related FDA regulatory improvement projects. 

Our approach consists of two main steps. In the first step, we collect and assemble FDA data related to 510(k) devices, and develop ML models capable of estimating each device's risk of recall based on the information available at the time of submission. Specifically, employing a text mining algorithm, we start by extracting information regarding both applicant devices and their corresponding predicates, including their characteristics and recall information. Through extensive discussions with our collaborator, we also conduct thorough data engineering to create variables that aid in predicting a recall event. Building on our analysis, we then train and test several ML models and pick the best one, which achieves a cross-validation Area Under the Curve (AUC) score of 0.78. Our findings using this ML model suggest that the number of recalls reported for predicate devices, along with variables relevant to the age of the predicates, hold significant predictive power for a recall event. This validates speculations and initial empirical findings in the recent literature but through an ML lens that is trained on large-scale data involving over 31,000 submissions produced by 12,000 manufacturers from over 65 countries. Interestingly, however, unlike these more known facts, our analysis indicates that the timing of recalls reported for predicates of an applicant device and the age of the predicates provide valuable information for predicting a recall event.

In the second step, utilizing the ML model for risk assessment, we proceed to develop a data-driven clearance policy that can assist the FDA in decision-making. To this end, we make use of an optimization approach that sets decision-making thresholds.
Our approach takes into account the existence of ranges within the risk spectrum where the predicted risk is not informative enough to make a recommendation, necessitating human expert attention. These thresholds are strategically determined to balance the precision of decision-making with the workload burden imposed on the FDA during the decision-making process. The primary computational challenge lies in enforcing a workload constraint (the rate of devices deferred for FDA evaluation), turning the optimization model into a non-convex optimization problem. To address this, we develop a nested search algorithm that provides an approximation by leveraging the structural properties of our problem. This algorithm efficiently identifies effective thresholds on predicted recall risks through an iterative solution procedure. We also emphasize that our policy is designed to be capable of providing additional information for applicant devices that require further evaluation by human experts (e.g., by an FDA committee).

We \textcolor{black}{investigate} the potential benefits of our policy compared to the current practice of the FDA using a unique large-scaled dataset that we assembled based on the public FDA datasets (Section~\ref{Empirical Results}). Our results \textcolor{black}{indicate} a 32.9\% improvement in the recall rate and a 40.5\% reduction in the FDA's workload.
Additionally, we \textcolor{black}{estimate} the potential cost savings from implementing our data-driven policy through estimating medical device replacement costs by medical specialties (Section~\ref{Policy Impact: Costs}). For the very first time, this is done by carefully determining the medical specialty for over 99\% of the 1,351 unique Healthcare Common Procedure Coding System (HCPCS) codes/descriptions for the CMS administrative claims data during 2013-2020. Our analysis indicates significant potential annual cost savings of 
\$1.7 billion from implementing our policy.

Our results provide valuable insights for the FDA and address main concerns about the current 510(k) process (Section~\ref{ManagerialInsights}). First, we observe that while there are some easy cases suitable for direct algorithmic decision-making, some other cases are difficult and require more in-depth evaluation and human judgment. This observation confirms our hypothesis that there is a need for a combined human-algorithm approach to integrate the FDA's expertise with quantitative evidence. Next, our results address the FDA’s concerns communicated in its recent solicitation announcements as well as its goal to establish best practices to evaluate the safety of medical devices (\citealt{FDABestPractices}). Specifically, we find that best practices should take advantage of the fact that the number of recall events for predicates, particularly recent recalls, is a significant predictor of recall risk for an applicant device. Furthermore, we observe that the age of the predicates is among other important predictors of recall risk. Finally, our results contribute to the recent call for feedback by the FDA on the opportunities and challenges of using artificial intelligence (AI) and ML in the development of drugs and medical devices (\citealt{FDAAILML}). Our work highlights the importance of crucial considerations in the context of utilizing AI/ML to enhance the evaluation of medical devices, and highlights the importance of employing a combined human-algorithm approach.

\subsection{Literature Review} \label{Literature Review}

Our paper contributes to \textcolor{black}{three} main bodies of research.
Below, we concentrate on the most closely related works.

\noindent \textbf{Medical Device Recalls.} There is a vast literature focusing on comparing the risk of recall for devices that received approval via the 510(k) pathway and those that received approval via premarket approval (see, e.g., \citealt{day2016analysis}, \citealt{connor2017medical}, \citealt{janetos2017overview}, \citealt{talati2018major}, and \citealt{dubin2021risk}). There are also a few empirical studies (see, e.g., \citealt{wowak2021influence} and \citealt{ball2018responding}) related to our work that examine factors influencing recall-related events of medical devices. Another related set of studies investigates the relationship between the characteristics of predicate medical devices and the recall events of 510(k) devices (\citealt{everhart2023association} and \citealt{kadakia2023use}). The study of \citealt{mukherjee2018product} presents a predictive model for the risk of recall using the adverse events occurring after the approval time. However, predictive models focusing on estimating the recall risk of an applicant device based on the information available at the time of submission remain scarce. In our study, we address this gap by developing a recall risk prediction model and recommending a clearance policy for the FDA, wherein the policy's recommendations are constrained by the available information at the time of the 510(k) application submission as well as the workload that can be imposed to the FDA for further evaluations by human experts prior to decision-making. Our contributions to this literature are twofold. First, we conduct a thorough examination of factors that prove useful in predicting the risk of recall, and design ML models that can benefit from the available information at the submission time. Second, we propose the first data-driven clearance policy that integrates the predictive power of ML models into the decision-making process in order to improve the FDA's 510(k) regulatory pathway. 

\noindent \textbf{Threshold-Based Decision-Making.} 
There are several existing methods to utilize risk predictions for decision-making by determining a single decision threshold, ranging from utility-based methods (see e.g., \citealt{jund2005methods}, \citealt{felder2014risk}, and \citealt{van2018alternative}) \textcolor{black}{to the receiver operating characteristic (ROC) method} (see, e.g.,
\citealt{hajian2013receiver} and \citealt{hong2021diagnostic}) and Bayesian decision theory (see, e.g., \citealt{sheppard2005bayesian} and \citealt{weise2006bayesian}). Our work is closest to the subset of literature focusing on three-way classifications with three categories of positive, negative, and undecided based on the evidence (see, e.g.,   \citealt{baram1998partial}, \citealt{si2017sequential}, \citealt{yao2010three}, \citealt{yao2016two}, 
\citealt{mozannar2020consistent},
\citealt{garcia2020data}). However, our work differs from these studies in two major ways. First, balancing workload plays a significant role in our work. In a three-way classification using two decision thresholds, setting the threshold conservatively often results in achieving a high performance measure in the positive and negative regions. However, it may lead to assigning many other instances to the undecided region which often will require attention from human experts. Consequently, this approach may not significantly reduce the workload in the system, though balancing the workload (speed) and quality of the decisions made is a crucial element in many systems (see, e.g., \citealt{saghafian2018workload}). 
Second, most of the existing methods for finding the decision thresholds without workload considerations are less complex and often can be solved using simple heuristics (\citealt{si2017sequential}), Bayesian rough sets (\citealt{yao2010three}, \citealt{yao2016two}), or a linear program (\citealt{garcia2020data}). In contrast, our methodology requires different techniques to handle the non-linearity in optimizing decision thresholds. Our solution technique is obtained by deriving structural properties of our optimization model and introducing a nested search approach.

\noindent \textbf{Human-in-the-Loop Approaches.} An expanding body of research suggests that ML models can outperform humans in making predictions across a wide range of domains (see, e.g., \citealt{liu2018mlbench}, \citealt{shen2019artificial}, \citealt{boloori2022understanding}, \citealt{ang2022using}). \textcolor{black}{Typically, there are two main motivations for adopting human-in-the-loop approaches. The first motivation is related to stakeholder trust.} While the state-of-the-art ML models exhibit impressive performance, in high-risk domains such as healthcare, there is reluctance to fully embrace automated systems and eliminate humans entirely from the loop due to the inherent distrust and lack of robustness (\citealt{american2019augmented}, \citealt{dean2025implementation}).

\textcolor{black}{The second motivation involves complementary strengths and performance gains.} Recent literature has highlighted the benefits of incorporating human judgment into the ML model deployment process, leading to performance gain. The idea of keeping humans in the loop has been implemented in various ways. There have been attempts to design interactive and active ML systems that continuously learn from humans (\citealt{wu2022survey}) or even go beyond human-in-loop mechanisms by systematically incorporating symbiotic learning (\citealt{muller2022ai}, \citealt{saghafian2023effective}). Other attempts include combining separate human and algorithm outputs (\citealt{blattberg1990database}, \citealt{goodwin2000correct}), introducing systems to elicit human judgment for prediction algorithms (\citealt{ibrahim2021eliciting}), and learning the human experts' intuition for risk prediction (\citealt{orfanoudaki2022algorithm}). Another approach to keeping humans in the loop is AI augmentation, where the main idea is to have AI systems work alongside humans and collaborate with them. This idea is different from automation that results in replacing humans with AI (\citealt{daugherty2018human}, \citealt{miller2018ai}). Our work provides further evidence on the latter approach, providing a regulatory policy that allows human experts to concentrate on complex cases with the assistance of AI/ML.

The remainder of the paper is organized as follows. Section~\ref{Setting and Data} provides a summary of our data collection and pre-processing steps. Section~\ref{Prediction} describes the development and evaluation of prediction models. Section~\ref{Policy} introduces our data-driven clearance policy, its analytical properties, and our solution methodology. Section~\ref{Empirical Results} provides our empirical results. Finally, we present managerial insights and concluding remarks in Section~\ref{ManagerialInsights}.

\section{Setting and Data} \label{Section2} \label{Setting and Data}
In this section, we start by briefly explaining our data collection and pre-processing steps. We then discuss various aspects of our data and provide a data summary.


\subsection{\textbf{Data Collection}} \label{DataCollection} 
\begin{figure}[tb]
\centerline{\includegraphics[scale=.35]{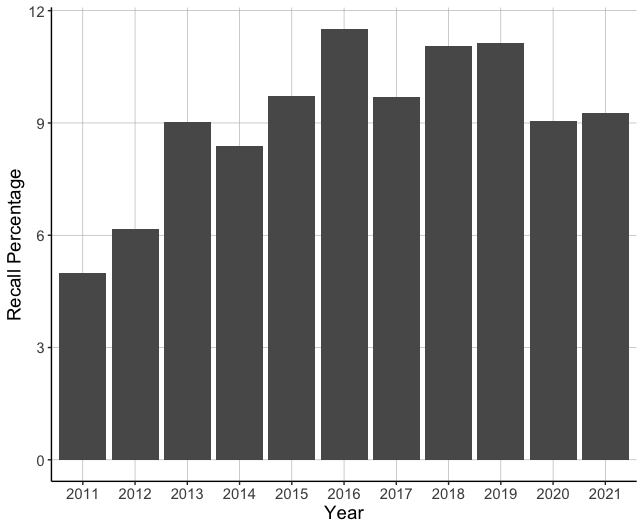}}
\caption{Recall percentage per year}
\label{fig1}
\end{figure}

We collected 510(k) applicant device submission data for the years 2008 to 2020 and FDA recall data for the years 2008 to 2021 from public FDA datasets. The 510(k) data includes the unique 510(k) number for each application, submission and clearance dates as well as device characteristics, such as medical specialty, product type, and device class. Figure~\ref{fig1} shows the recall percentage from 2011 to 2021.
The FDA recall data contains the 510(k) number associated with the recalled devices, event dates, and severity of recalls. 
The FDA categorizes recall events into three major classes based on the relative degree of health hazard that can be posed to patients. A Class 1 (severe) recall  is a situation with a reasonable probability of serious adverse health consequences or death. A Class 2 (moderate) recall is a situation where temporary or medically reversible adverse health consequences are probable. A Class 3 (mild) recall is a situation that is not likely to cause adverse health consequences. 

In the publicly available datasets, there is no direct link between an applicant device and its recall information. For each applicant device, we therefore directly investigated whether it had any prior recalls and identified the recall class when a recall event was reported. \textcolor{black}{A significant data limitation is the lack of structured data linking applicants to their predicate devices. To address this}, we developed a text mining algorithm to identify the predicate devices listed by the manufacturers for each cleared 510(k) applicant device using the summary documents associated with each submitted application. The text mining algorithm is described in detail in Appendix \ref{EC:TextMining}. \textcolor{black}{Our resulting data included the applicant devices for which the algorithm identified at least one predicate device. Approximately 3\% of devices, mainly early scanned summaries, were excluded. The excluded devices are (i) concentrated in 2008–2011, a period that predates FDA’s mandatory eCopy rule (\citealt{FDAeCopy}), (ii) spread proportionally across different medical specialties with exclusion rates ranging from 0.9\% to 8.2\%, and (iii) similarly spread across manufacturing countries, ranging from 0.6\% to 5.5\%.} We also extracted device characteristics and recall information for all the predicate devices and added them to our dataset.

Our dataset, assembled based on the publicly available FDA 510(k) records, contains only those devices that received market authorization. However, FDA does not reveal information on devices that were deemed not substantially equivalent (NSE) and thus rejected, primarily due to confidentiality restrictions (21 CFR 807.95). According to internal analyses from the Center for Devices and Radiological Health (CDRH), approximately 5\% of all 510(k) submissions result in NSE decisions (\citealt{CDRH}). While this relatively small fraction of rejections may mitigate concerns about selection bias, it does not eliminate them. In Section~\ref{Empirical Results}, we discuss a robustness check conducted to address this limitation in our dataset. \textcolor{black}{In Appendix~\ref{EC:SelectionBias}, we discuss in detail why only a fraction of NSE cases would be usable for performance evaluation rather than model training, and provide a robustness check to address this limitation.}

\subsection{\textcolor{black}{\textbf{Summary of the Study Sample}}} 
Our primary outcome is a \textit{binary recall event} indicating whether the applicant device had at least one recall between its FDA clearance date and the end of our study period. \textcolor{black}{We constructed a comprehensive set of predictive features capturing applicant characteristics, predicate characteristics, and predicate recall history. A detailed description of all variables and their construction is provided in Appendix \ref{EC:VariableDefinitions}. Tables \ref{table100} and \ref{table200} provide summary statistics for the study sample.}

\begin{table} [tbh]
\centering \footnotesize
\caption{Summary of the characteristics of the 510(k) submissions in the study sample}
\label{table100}
\begin{tblr}{
  width = \linewidth,
  colspec = {X[3,l] X[1,c] X[1,c] X[1,c]}, 
  hline{1-2} = {-}{}, 
  hline{28} = {-}{}, 
}
    & {\textbf{No Recall }\\\textbf{(N=28,611)}} & {\textbf{Recall }\\\textbf{ (N=3,293)}} & \textcolor{black}{{\textbf{P-value}} } \\
    
    Medical Specialty (Top and Low Three) & & & $<\;$0.001 \\
    ~ ~ Radiology (RA) & 3,532 (12.3\%) & 664 (20.2\%) & \\
    ~ ~ Orthopedic (OR) & 5,551 (19.4\%) & 630 (19.1\%) & \\
    ~ ~ Cardiovascular (CV) & 3,966 (13.9\%) & 483 (14.7\%) & \\
    ~ ~ Ear, Nose, Throat (EN) & 303 (1.1\%) & 24 (0.7\%) & \\
    ~ ~ Clinical Toxicology (TX) & 249 (0.9\%) & 20 (0.6\%) & \\
    ~ ~ Physical Medicine (PM) & 555 (1.9\%) & 13 (0.4\%) & \\
    
    Device Class & & & $<\;$0.001 \\
    ~ ~ I & 934 (3.3\%) & 26 (0.8\%) & \\
    ~ ~ II & 27,677 (96.7\%) & 3,267 (99.2\%) & \\
    
    Country Code (Top and Low Three) & & & $<\;$0.001 \\
    ~ ~ United States (US) & 19,978 (69.8\%) & 2,679 (81.4\%) & \\
    ~ ~ Other & 1,886 (6.6\%) & 176 (5.3\%) & \\
    ~ ~ Germany (DE) & 838 (2.9\%) & 90 (2.7\%) & \\
    ~ ~ Switzerland (CH) & 392 (1.4\%) & 31 (0.9\%) & \\
    ~ ~ Taiwan (TW) & 625 (2.2\%) & 17 (0.5\%) & \\
    ~ ~ Korea (KR) & 838 (2.9\%) & 15 (0.5\%) & \\
    
    Product Code (Top and Low Three) & & & $<\;$0.001 \\
    ~ ~ Other\_OR & 3,180 (11.1\%) & 366 (11.1\%) & \\
    ~ ~ Other\_CV & 3,016 (10.5\%) & 344 (10.4\%) & \\
    ~ ~ Other\_RA & 1,493 (5.2\%) & 238 (7.2\%) & \\
    ~ ~ IYN & 500 (1.7\%) & 86 (2.6\%) & \\
    ~ ~ JAK & 206 (0.7\%) & 82 (2.5\%) & \\
    ~ ~ Other\_AN (\textcolor[rgb]{0.18,0.161,0.145}{Anesthesiology}) & 817 (2.9\%) & 79 (2.4\%) & \\
    
    Implantable & 7,950 (27.8\%) & 868 (26.4\%) & \;\;\;0.081 \\
    
    Life Sustaining/Supporting & 703 (2.5\%) & 222 (6.7\%) & $<\;$0.001 \\

\SetCell[c=4]{l}
    {\parbox{\dimexpr\textwidth-12pt}{
        \footnotesize 
        Note: Data are presented as number (percentage). 
        \textcolor{black}{P-values are calculated using the Chi-squared test for difference in proportions.}
    }} & & & \\
\end{tblr}
\end{table}

\begin{table} [tbh]
\centering \footnotesize
\caption{Summary of the predicates' characteristics and recall information of the 510(k) submissions in the study sample}
\label{table200}
\begin{tblr}{
  width = \linewidth,
  colspec = {X[3,l] X[1,c] X[1,c] X[1,c]}, 
  hline{1-2} = {-}{}, 
  hline{16} = {-}{}, 
}
    & {\textbf{No Recall }\\\textbf{ (N=28,611)}} & {\textbf{Recall }\\\textbf{ (N=3,293)}} & \textcolor{black}{{\textbf{P-value}}} \\
    
    \textit{Predicate Devices’ Characteristics} & & & \\
    Num. of Predicate & 2.36 (2.29) & 2.69 (3.18) & $<\;$0.001 \\
    Prop. of Unmatched Medical Specialties & 0.077 (0.225) & 0.070 (0.212) & \;\;\;0.059 \\
    Prop. of Unmatched Product Code & 0.194 (0.308) & 0.210 (0.313) & \;\;\;0.004 \\
    Predicate Median Age & 4.09 (3.62) & 3.84 (3.30) & $<\;$0.001 \\
    Predicate Newest Age & 3.85 (4.72) & 3.32 (4.06) & $<\;$0.001 \\
    Predicate Oldest Age & 7.44 (6.61) & 7.37 (6.55) & \;\;\;0.562 \\
    
    & & & \\
    
    \textit{Recall Information} & & & \\
    Number of Class 1 & 0.014 (0.202) & 0.036 (0.315) & $<\;$0.001 \\
    Number of Class 2 & 0.387 (1.84) & 1.22 (3.95) & $<\;$0.001 \\
    Number of Class 3 & 0.009 (0.103) & 0.019 (0.175) & $<\;$0.001 \\
    Variance of Recalls & 0.040 (0.896) & 0.163 (1.98) & $<\;$0.001 \\
    Weighted Recall Score & 0.109 (0.264) & 0.262 (0.373) & $<\;$0.001 \\

\SetCell[c=4]{l}
    {\parbox{\dimexpr\textwidth-12pt}{
        \footnotesize
        Note: Data are presented as mean (standard deviation).
        \textcolor{black}{P-values are calculated using the t-test for difference in means.}
    }} & & & \\
\end{tblr}
\end{table}

\section{Machine Learning Models for Predicting Recall Events} \label{Prediction}
\textcolor{black}{We now describe the training and evaluation of ML models for predicting recall events.} Our ML models are trained on our data described in the previous section which contains information on a variety of medical devices with different characteristics in order to predict recall events of unseen future applicant devices. We consider various ML models and compare their performance to pick the best one. The models we consider are regularized logistic regression (Log Reg) using both Lasso and Ridge type penalties, decision tree, random forest, and gradient boosting. \textcolor{black}{All models were trained in Python using the scikit-learn (sklearn) library.}

To train and tune our ML models, we split the data into training (70\%) and testing (30\%) sets using stratified sampling to preserve the original class distribution. While a temporal split could potentially be utilized to simulate real-world deployment, we opted for a non-temporal split with stratified sampling because the low and fluctuating prevalence of recall events in our dataset required a split that ensures stable and representative class proportions across both sets. Furthermore, this approach avoids the limitations of the rolling horizon cross-validation required in a temporal split, where models trained on early, sparse historical data may lack sufficient stability for robust hyperparameter tuning.

We then employ a stratified cross-validation approach on the training set to fine-tune hyperparameter values of each ML model. Specifically, we perform 10-fold cross-validation repeated 10 times on the training set to evaluate the average area under the receiver operating characteristic curve (AUC) for each hyperparameter combination of each model. This robust approach helps reduce overfitting by selecting hyperparameter values that generalize well to unseen data. In particular, for logistic regression models, we optimize the regularization parameter. Similarly, for decision trees, we fine-tune the depth and the minimum number of samples required to split a node. For ensemble methods, we optimize two key parameters for each model: the number of trees and the maximum depth for random forests, and the number of trees and the learning rate for gradient boosting. We select the set of hyperparameter values that maximize this cross-validation AUC (CV-AUC) as the optimal configuration for each model. The implementation details of the ML models and cross-validation procedures are outlined in Appendix \ref{Appendix-implementation}.

\subsection{\textbf{Performance Evaluation of the Machine Learning Models}} \label{predictive model results}
The CV-AUC for the best parameter set of each model, along with the corresponding 95\% confidence interval (CI) derived from the quantiles of the repeated cross-validation runs, is presented in Table~\ref{table-AUC}. We find that all models except the decision tree model attain relatively similar performance in terms of the CV-AUC metric, ranging from 0.77 to 0.78. 
To calculate the AUC on the testing set, we trained each model on the entire training set using the best hyperparameter values selected during cross-validation. The out-of-sample AUC (OOS-AUC) of each model, as well as the corresponding 95\% CI generated using bootstrapping on the testing set, is also presented in Table~\ref{table-AUC}. As shown in the table, the final OOS-AUC scores for the top-performing models were identical (0.76), confirming their comparable predictive power on unseen data and validating our cross-validation results. We note that the decision tree model underperforms compared to the other models by a slight margin. Based on our further investigation, we find that its relatively weaker performance is not primarily due to overfitting, but is more likely explained by the structural limitations of the model. While a single tree attempts to model continuous risk factors using coarse, orthogonal splits, the ensemble methods and logistic regression are better equipped to approximate the smooth, monotonic relationships likely present in the data.

\color{black}
\textcolor{black}{To further refine our ML models, we explored several common strategies. First, to mitigate class imbalance, we applied the synthetic minority over-sampling technique (SMOTE) to balance the training data. We also used class weighting technique, which adjusts the loss function to penalize misclassification of minority class instances more heavily. However, neither of these strategies led to meaningful improvements in model performance in terms of AUC. Second, since three of our predictors (Medical Specialty, Product Code, and Country Code) contain a large number of unique values, we also tested ML models better suited for handling high-cardinality categorical variables in tree-based ML models. Specifically, we used the CatBoost and LightGBM packages, but the trained models did not outperform our models listed in Table~\ref{table-AUC}. Finally, devices cleared by the FDA close to the end of our study period may have had less time to experience a recall than those cleared earlier. This could create a censoring effect for the most recent applicant devices. We employed two robustness checks to assess the potential impact of this censoring. The results suggest that our selected model is not impacted and provides a reliable method of recall risk prediction (see Appendix \ref{Appendix-Censoring}).} 


\begin{table}[b]
\centering \small
\caption{Performance comparison of ML models}
\label{table-AUC}
\begin{tabular}{@{}l c c@{}} 
\toprule
\textbf{Model} & \textbf{CV-AUC (95\% CI)} & \textbf{OOS-AUC (95\% CI)} \\
\midrule
Log Reg with Ridge penalty & 0.77 (0.73, 0.80) & 0.76 (0.75, 0.77) \\
Log Reg with Lasso penalty & 0.77 (0.73, 0.80) & 0.76 (0.75, 0.77) \\
Decision tree              & 0.75 (0.71, 0.78) & 0.73 (0.71, 0.74) \\
Random forest              & 0.77 (0.73, 0.80) & 0.76 (0.75, 0.77) \\
Gradient boosting          & 0.78 (0.74, 0.81) & 0.76 (0.75, 0.78) \\
\bottomrule
\end{tabular}
\end{table}

\begin{figure}[tb]
\centerline{\includegraphics[scale=.45]{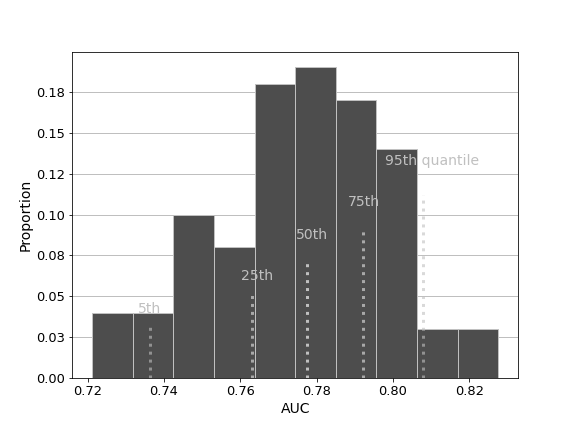}}
\caption{Distribution of CV-AUC scores for the selected model (gradient boosting)}
\label{fig3}
\end{figure}

Among the different models, we ultimately chose gradient boosting for two reasons. First, while its predictive performance was  comparable to random forest and logistic regression models as evidenced by the 95\% CV-AUC CIs, it consistently demonstrated the highest CV-AUC. Second, and more importantly, its capacity to capture more complex relationships (e.g., potential interactions) makes it more adaptable should the FDA incorporate additional features in the future. The distribution of CV-AUC scores for gradient boosting is depicted in Figure~\ref{fig3}, where the vertical lines correspond to different quantiles. \textcolor{black}{Ultimately, however, the final model selection rests with the FDA, depending on their specific prioritization of predictive performance versus interpretability.}

\begin{figure}[tb]
    \centering
    \includegraphics[scale=0.5]{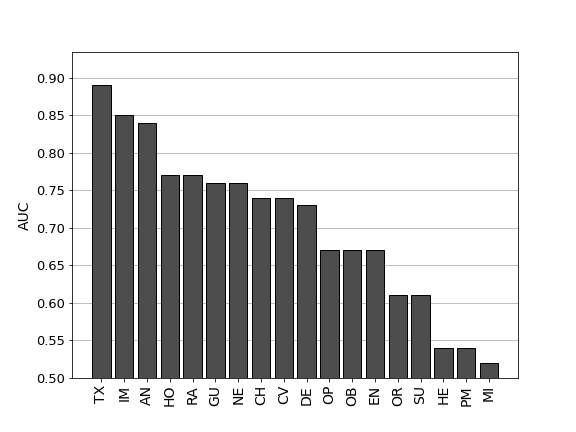}
    \caption{OOS-AUC across all specialties}
    \label{fig4}
  \begin{minipage}{14cm}
  \vspace{5pt}
    \footnotesize Note: TX = Clinical Toxicology; AN = Anesthesiology; IM = Immunology; RA = Radiology; \\ HO = General Hospital; NE = Neurology; CV = Cardiovascular; GU = Gastroenterology \& Urology; CH = Clinical Chemistry; EN = Ear, Nose, \& Throat;  OB = Obstetrics/Gynecology; \\ OP = Ophthalmic; DE = Dental; SU = General \& Plastic Surgery; OR = Orthopedic; \\ HE = Hematology; MI = Microbiology; PM = Physical Medicine (PM)
  \end{minipage}
\end{figure}

The AUC results reported in Table~\ref{table-AUC} are calculated based on the overall performance of models across all medical specialties. In practice, each medical specialty has a separate committee for evaluating the applicant devices assigned to that medical specialty. The 510(k) process can be viewed as a general rule, but each committee's evaluation criteria may vary. We evaluate the predictive power of gradient boosting separately for each medical specialty. \textcolor{black}{Figure~\ref{fig4} shows the OOS-AUC per specialty. As can be seen, the model has the best performance in terms of the AUC metric for applicant devices of Clinical Toxicology (TX), Immunology (IM), and Anesthesiology (AN). On the other hand, Hematology (HE), Physical Medicine (PM), and Microbiology (MI) are three medical specialties for which the model has the worst performance.}

\begin{figure}[tb]
\centerline{\includegraphics[scale=.58]{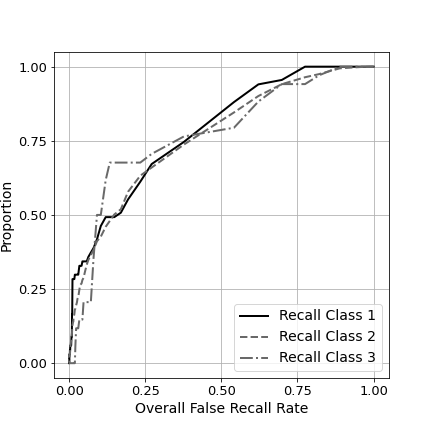}}
\caption{Proportion of recall classes correctly identified versus the overall false recall rate}
\label{fig5}
\end{figure}

When it comes to a recall event, the severity of the recall can be relatively identified by the FDA's classification of recalls. Although there is value in detecting each recall class, detecting a high-severity recall class has the highest priority due to its life-threatening nature. We evaluate our gradient boosting model on its ability to detect different classes of recalls. Figure~\ref{fig5} plots the proportion of recalls correctly identified per each recall class versus the overall false recall rate. Overall, the model has better predictive power for Class 1 recalls, followed by Class 2 and Class 3. \textcolor{black}{In particular, it achieves an OOS-AUC of 0.78, 0.76, and 0.75 in detecting recall Classes 1, 2, and 3, respectively.} This is a desirable performance since Class 1 has the highest severity, followed by Class 2 and then Class 3. 

\begin{figure}[tb]
 \centering
\centerline{\includegraphics[scale=.7]{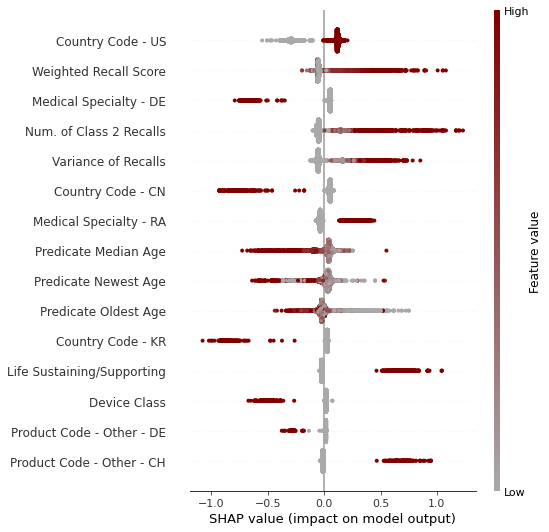}}
\caption{Top 15 predictors of recall risk}
\label{fig-SHAP-gb}
 \begin{minipage}{17cm}
  \vspace{5pt}
    \footnotesize Note: US = United States; DE = Germany; CN = China; RA = Radiology; KR = Korea; CH = Clinical Chemistry 
  \end{minipage}
\end{figure}

Lastly, we investigate the most important variables and their impact on predicting recall risk using our gradient boosting model. We use the \textcolor{black}{Shapley} additive explanations (SHAP) method (\citealt{lundberg2017unified}, \citealt{lundberg2020local}), which leverages a game theory approach to compute the contribution of variables to a predicted value in an additive form. We compute the contribution of each variable to the predicted recall risk.
Figure~\ref{fig-SHAP-gb} highlights the 15 most important variables and their impact on the predicted recall risk. They are ordered by decreasing significance. In Figure~\ref{fig-SHAP-gb}, each point represents a variable's contribution to the prediction. The value of each contribution is depicted by a color gradient from grey to red, where grey indicates low values and red indicates high values. Our results indicate that variables related to applicant device characteristics, the age of predicate devices, and the recall status of predicate devices are important predictors of recall risk. \textit{Weighted Recall Score} accounts for the timing of predicate recall events. We observe that a higher value of this variable is associated with a higher recall risk. We also find that \textit{Num. of Class 2 Recalls} is highly predictive of the recall risk. These observations not only indicate that the number of predicate recall events matters, but also highlight the importance of their timing, an aspect that has been overlooked in the literature. This supports our hypothesis that paying attention to predicate devices with recent recalls can go a long way in raising red flags for an applicant device. In particular, a recall event that has happened long ago may have been fully addressed, but some uncertainties might remain unresolved for more recent recalls. Our results also highlight that the \textit{Variance of Recalls} is predictive of the recall risk.

With regard to the age of predicates, we observe that a higher value of \textit{Predicate Newest Age} is associated with a lower recall risk. 
This may be because newer predicates might not have undergone extensive usage, which can reveal potential issues over time. We observe that high values of \textit{Predicate Oldest Age} can lead to a low recall risk. This aligns with our earlier observation that older predicates can indicate safety, as they have a history of safe and effective use over time and may be the gold standard for patient care. With regard to \textit{Predicate Median Age}, we observe that, overall, a higher value of this predictor is associated with a lower recall risk. Altogether, these findings highlight a learning curve effect. That is, if a predicate device has been on the market only briefly, latent problems may remain undiscovered that may lead to a higher risk of future recalls.

Among the variables corresponding to the characteristics of applicant devices, medical specialties, country codes, and product codes are highly predictive of the recall risk. The results are consistent with prior literature indicating some heterogeneous effects of these indicators on the risk of recall.
\textcolor{black}{Regarding the high predictive importance of the ``US" country code, we hypothesize that this may reflect regulatory and liability dynamics rather than device quality alone. U.S. manufacturers are subject to more frequent facility inspections and operate within a distinct liability landscape (\citealt{FDAinspection}), factors that likely increase the rate of voluntary reporting and defect identification.} Furthermore, our results show that \textit{Life Sustaining/Supporting} is of significant importance. This variable quantifies the risk level associated with an applicant device. Riskier applicant devices are generally under stringent market scrutiny and are more likely to be recalled. \textcolor{black}{
Finally, with respect to the similarity between an applicant device and its predicate devices, we observe that neither \textit{Prop. of Unmatched Specialties} nor \textit{Prop. of Unmatched Product Codes} is among the top 15 most informative variables. This is notable given that device dissimilarity has been historically deemed to contribute to failures like metal-on-metal hip implants (\citealt{ardaugh2013510}). Our finding,  along with recent work in the literature (\citealt{everhart2023association}), suggests that broad categories may fail to capture the granular design differences that drive risk, or that such high-level dissimilarity is not inherently predictive of recall.}

To validate our key findings regarding important predictors of recall risk, we examine the logistic regression model with Lasso, which achieves performance comparable to our selected model. Specifically, 
we evaluate the $p$-values from a refitted logistic regression model that includes only the variables selected by Log Reg with Lasso (see Appendix \ref{Appendix-VarSig}). The results confirm that our findings are robust, and the interpretability provided by the SHAP analysis is highly consistent with the statistical inference offered by linear models. \textcolor{black}{Regarding the variables \textit{Predicate Median Age}, \textit{Predicate Oldest}, and \textit{Variance of Recalls}, although they contribute to improved predictive performance in the selected model, they were not selected as significant predictors by the Lasso logistic regression model. These results suggest that their association with recall risk should be interpreted with caution. }

\color{black}
\section{Design of a Data-Driven Human-Algorithm Decision Support Tool} \label{Policy}
In this section, we make use of our best predictive model discussed in the previous section, and introduce a data-driven clearance policy that balances increased safety and expeditious evaluation of medical devices. The FDA has designated committees that evaluate devices seeking clearance through the 510(k) pathway. The policy we propose is based on a human-algorithm approach designed to improve and facilitate the 510(k) approval pathway by reducing the risk of potential recalls and the workload of the FDA’s designated committees. In Section~\ref{Empirical Results}, we perform various empirical investigations to estimate the impact of our proposed policy and further guide the FDA in implementing it.

A typical approach in designing a policy to recommend suitable actions based on an ML model's outputs is to estimate the utility of possible outcomes (recall risk), and impose a fixed threshold on the predicted risk to maximize the utility. 
Unfortunately, relying solely on a single threshold fails to identify specific regions where risk estimation models perform poorly. In other words, it overlooks the range of risk estimate values where false-positive and false-negative rates are high, which are cases where diagnostic decisions should be delegated to human experts and approached with caution. 

We develop an advanced clearance policy by harnessing the power of an ML model. 
However, in cases where the predicted risk for an applicant device may not be informative enough, our policy has the capability to offer supplementary information for assessment by human experts without presenting a direct recommendation. The evaluation of such intricate cases is delegated to the designated FDA committee, leveraging their expertise to arrive at more informed judgments. 

\subsection{A Data-Driven Advanced Clearance Policy} \label{policy description}
Our policy has two main components: (1) an ML predictor to estimate the recall risk of a 510(k) device based on the information available to the FDA upon submission by the manufacturer, and (2) an optimization approach that determines whether an applicant device can be accepted/rejected or if it should be deferred and more elaborately evaluated by an FDA-assigned committee. For deferred devices, the estimated risk can be provided to the FDA-assigned committee as additional information to assist the committee in making better judgments. 
For devices recommended for acceptance or rejection by our policy, we envision a minimal level of human review to ensure that the final decision remains under FDA's oversight. This modest review could involve a brief verification process that requires only a small fraction of the effort typically needed for a full FDA evaluation. 
By combining these main core elements, our policy streamlines the clearance process, ensuring that devices are assessed with greater efficiency and accuracy while there are enough resources for in-depth evaluation of deferred devices that require more attention.

\begin{figure}[tb]
\centerline{\includegraphics[scale=.6]{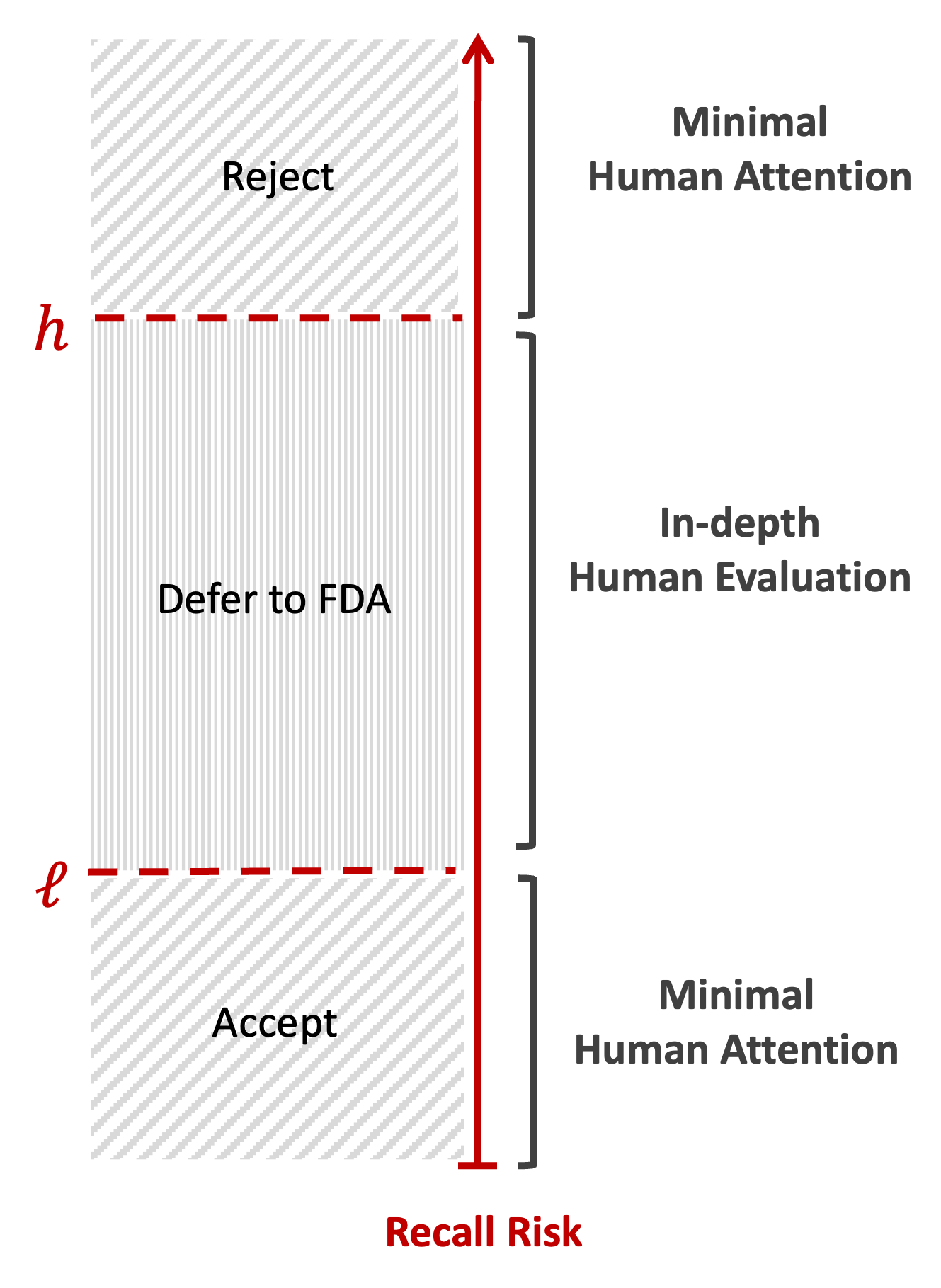}}
\caption{Schematic view of the proposed clearance policy}
\label{fig7}
\end{figure}

In our policy, the evaluation occurs through two main phases, as depicted in Figure~\ref{fig7}. In the initial phase, the ML model discussed in Section~\ref{Prediction} assesses the recall risk of an applicant device based on the available information provided during the submission. If the predicted risk by the ML model is lower than an optimized low threshold ($\ell$), the policy recommends accepting the device. If, however, the predicted risk exceeds the high threshold ($h$), the device is recommended to be rejected. The policy optimizes the values of the low and high thresholds to minimize the rates of \textit{acceptance of unsafe devices} and \textit{rejection of safe devices}, while still aligning with the preferences of decision-makers. That is, it ensures that (a) rates of \textit{rejection of unsafe devices} and \textit{acceptance of safe devices} are at least greater than values specified as desired and (b) the workload imposed on the FDA for more elaborate evaluation of deferred devices does not exceed a preferred level.

As mentioned, the first core element of our policy is predicting the recall risk for an applicant device given the information available upon application submission. In Section~\ref{predictive model results}, we discussed how we have developed and trained a well-performing ML model using our dataset. Given a vector of inputs containing information on an applicant device and its corresponding predicate devices, the model generates a probability of recall. Let $f: X \rightarrow (0,1)$ be the functional representation of the model, which maps the vector of information on an applicant device and its predicates denoted by $X$ to a probability value in $(0,1)$. \textcolor{black}{This function approximates the conditional probability $\mathbb{P}(Y=1|X)$.} The second core element of our policy contains developing a constraint optimization model to identify the low and high thresholds. \textcolor{black}{Let $\delta^{+} = f(X)|Y=1$ be the recall risk estimated for an applicant device which has been recalled according to data ($Y=1$). Similarly, let $\delta^{-} = f(X)|Y=0$ be the recall risk estimated for an applicant device with no recall in data ($Y=0$). Note that both $\delta^{+}$ and $\delta^{-}$ are random variables as they are functions of $X$.} Using this notation, we next describe the logic behind the optimization model that forms the second core element of our proposed policy.

A well-designed and effective policy is expected to yield low rates of acceptance of unsafe devices $(\mathbb{P}(\delta^{+} \le \ell))$ and rejection of safe devices $(\mathbb{P}(\delta^{-} \ge h))$. However, a low value for those measures may yield low rates of rejection of unsafe devices $(\mathbb{P}(\delta^{+} \ge h))$ and acceptance of safe devices $(\mathbb{P}(\delta^{-} \le \ell))$. Also, a low workload $(\mathbb{P}(\ell < f(X) < h))$ for the FDA’s committees is desired, so only a low ratio of applicant devices should be deferred and judged by them. Due to trade-offs between these measures, identifying the values of low and high risk thresholds ($\ell$ and $h$, respectively) is challenging. We develop a non-linear optimization model to solve this challenging problem: 

\noindent \textcolor{black}{\textbf{Primary Problem} }
\begin{subequations}
\label{opt:main_model}
\begin{align} \min_{\ell,h}   &\;\; \lambda \; \mathbb{P}(\delta^{+} \le \ell) + (1-\lambda) \; \mathbb{P}(\delta^{-} \ge h)  \label{cc:1}\\
\text{s.t.}&\;\; \mathbb{P}(\delta^{+} \ge h) \; \ge \; \xi^{ru} \label{cc:2} \\  
&\;\;  \mathbb{P}(\delta^{-} \le \ell) \; \ge \; \xi^{as} \label{cc:3} \\ 
&\;\; \mathbb{P}(\ell < f(X) < h)  \; \le \; \rho \label{cc:4}\\ 
&\;\; 0 \; \le \; \ell \; \le \; h \; \le \; 1.  \label{cc:5}
\end{align}
\end{subequations}
\color{black}
The objective is to minimize the weighted sum of rates of acceptance of unsafe devices and rejection of safe devices. 
Although rejecting a safe device may cause a delay in the clearance process for the manufacturer, it is relatively less critical than accepting an unsafe device considering the availability of substantially equivalent devices in the market. However, our framework is general and allows specifying relative importance considerations via the weight $\lambda\in (0,1)$ in $\eqref{cc:1}$.

Constraint $\eqref{cc:2}$ ensures that the rate of rejection of unsafe devices is greater than a threshold ($\xi^{ru}$). Constraint $\eqref{cc:3}$ requires that the rate of acceptance of safe devices is greater than a threshold ($\xi^{as}$). Finally, constraint $\eqref{cc:4}$ mandates that the workload (measured via the rate of deferred devices) is less than a desired threshold ($\rho$). The modeling parameters $\lambda$, $\xi^{ru}$, $\xi^{as}$, and $\rho$ are specified based on the preference of the decision-maker. They may vary from year to year, depending on the FDA's resources and other factors. 

\textcolor{black}{We note that there is a trade-off between the rate of acceptance of unsafe devices (first term in the objective function) and the rate of acceptance of safe devices \eqref{cc:3}. Similarly, there is a trade-off between the rate of rejection of safe devices (second term in the objective function) and the rate of rejection of unsafe devices \eqref{cc:2}. The existence of these two constraints imposes minimum performance guarantees. The lack of these constraints, along with the absence of significant limitations on the FDA's workload, results in a policy that defers almost all devices to minimize the objective function.}

\subsection{Structural Properties} \label{Structural Properties}
In this section, we derive important structural properties of the \textcolor{black}{Primary Problem}, allowing us to find a closed-form solution in many instances and develop an efficient algorithm to find near-optimal results in other instances.

In the optimization problem, computation of low and high thresholds are tangled due to the presence of the FDA's workload constraint, along with the additional requirement that the low threshold must be less than the high threshold. To gain a deeper understanding of the underlying structural properties, we now focus on a relaxed version of the \textcolor{black}{Primary Problem}.

\noindent  \textcolor{black}{\textbf{Relaxed Problem}. This problem is defined by removing the constraint regarding the FDA's workload (constraint \eqref{cc:4}) from the Primary Problem.}

In the Relaxed Problem, there is an interplay between different components. Specifically, if the rate of rejection for unsafe devices is desired to be high,
there would be a corresponding escalation in the rate of rejection for safe devices, resulting in an increase in the second term of the objective function. This interplay stems from the fact that a more stringent rejection criterion for unsafe devices inherently involves an elevated level of algorithmic scrutiny, leading to more rejection of safe devices as well.

Conversely, when the rate of acceptance of safe devices is desired to be high,
it results in an increase in the rate of acceptance of unsafe devices, yielding an increase in the first term of the objective function. This connection stems from the fact that a higher acceptance rate for safe devices implies a looser algorithmic screening process, inadvertently leading to a higher acceptance rate for potentially unsafe devices.

To solve the Relaxed Problem, we introduce an auxiliary problem for which we can derive a closed-form solution. We then demonstrate how this auxiliary problem is in essence equivalent to the relaxed problem. We define the auxiliary problem as follows: 

\noindent \textbf{Auxiliary Problem} 
\begin{subequations}
\label{opt:aux_problem}
\begin{align} 
\max_{\ell,h}   &\; - \theta \; \ell + (1-\theta) \; h \label{c:1} \\
\text{s.t.}& \;\; h \; \le \; h(\xi^{ru}) \label{cc:13}\\
&\;\; \ell \; \ge \; \ell(\xi^{as}) \label{cc:14}\\
&\;\; 0 \; \le \; \ell \; \le \; h \; \le \; 1, \label{cc:15}
\end{align}
\end{subequations}
where $\theta \in (0,1)$. Also, $h(\xi^{ru}) \; = \; \sup \left\{h \in [0,1]: \; \mathbb{P}(\delta^{+} \ge h) \; \ge \; \xi^{ru}  \right\}$ denotes the highest value of $h$ for which the true positive rate is greater than or equal to $\xi^{ru}$, and $\ell(\xi^{as}) \; = \; \inf \left\{\ell \in [0,1]: \; \mathbb{P}(\delta^{-} \le \ell) \; \ge \; \xi^{as}  \right\}$ denotes the smallest value of $\ell$ for which the true negative rate is greater than or equal to $\xi^{as}$.
\color{black}

We next derive a closed-form solution for the Auxiliary Problem.

\begin{proposition}[{\bf Closed-Form Solution}]\label{prop1} For any $\theta \in (0,1)$ and a pair of $h(\xi^{ru})$ and $\ell(\xi^{as})$ in the Auxiliary Problem, we have:
\begin{enumerate} [label=(\alph*)]
\renewcommand\labelenumi{\normalfont(\alph{enumi})}
    \item if $h(\xi^{ru}) \;>\; \ell(\xi^{as})$, then $(\ell(\xi^{as}), h(\xi^{ru}))$ is the unique optimal solution,
    \item if $h(\xi^{ru}) \;<\; \ell(\xi^{as})$, then the problem is infeasible, 
    \item if $h(\xi^{ru}) \;=\; \ell(\xi^{as})$, then $\ell \;=\; h \;=\; h(\xi^{ru}) \;=\; \ell(\xi^{as})$ is the single threshold optimal solution.
\end{enumerate}
\end{proposition}

This proposition establishes that when the Auxiliary Problem is feasible, there exists a closed-form optimal solution. Additionally, it indicates that the Auxiliary Problem is decomposable and its optimal solution is independent of the value of the parameter $\theta$, which is not generally true for the \textcolor{black}{Primary Problem}.

Lastly, we establish a connection between the Relaxed Problem and the Auxiliary Problem by demonstrating that the solution of the Auxiliary Problem can be used to solve the Relaxed Problem.

\begin{theorem}[{\bf Connecting the Relaxed and Auxiliary Problems}]\label{thm1} For any $\theta \in (0,1)$ and $\lambda \in (0,1)$, and a pair of $h(\xi^{ru})$ and $\ell(\xi^{as})$, we have:
\begin{enumerate} [label=(\alph*)]
\renewcommand\labelenumi{\normalfont(\alph{enumi})}
    \item if $h(\xi^{ru}) \;>\; \ell(\xi^{as})$, then the two-threshold optimal solution of the Auxiliary Problem is optimal in the Relaxed Problem,
    \item if $h(\xi^{ru}) \;<\; \ell(\xi^{as})$, then both problems are infeasible, 
    \item if $h(\xi^{ru}) \;=\; \ell(\xi^{as})$, then the single threshold optimal solution of the Auxiliary Problem is optimal in the Relaxed Problem.
\end{enumerate}
\end{theorem}

\textcolor{black}{This theorem establishes that the optimal solution to the Relaxed Problem is independent of the preference parameter $\lambda$. This occurs because, in the absence of the FDA's workload constraint, the optimal thresholds are dictated entirely by the constraints \eqref{cc:2} and \eqref{cc:3}, which impose an upper bound for $h$ and a lower bound for $\ell$, respectively. Hence, the Relaxed Problem can be decomposed and its optimal solution is independent of the value of $\lambda$.}

\subsection{Nested Search} \label{nested search}
\textcolor{black}{While the structural properties in Section~\ref{Structural Properties} provide a closed-form solution when the workload constraint is non-binding, we also need to find a solution to the \textcolor{black}{Primary Problem} in Section~\ref{policy description} when the workload constraint is binding. This case is challenging because when the workload constraint is binding, it couples the computations of the low and high thresholds.} In this section, we leverage the structural properties of our problem and develop a nested search approach to find thresholds $\ell$ and $h$. 

We begin by reformulating the workload constraint \eqref{cc:4}. Let $N^+$ and $N^-$ denote the total number of unsafe and safe devices in the training data, respectively. Consequently, the total size of the training data is $N= N^+ + N^-$, and the proportion of unsafe devices is $q= N^+/N$. Then, we have:
\begin{align*}
     \mathbb{P}(\ell < f(X) < h)  &=  1- \left(  \mathbb{P}( f(X) 
    < \ell)  + \mathbb{P}(f(X) > h)   \right),
\end{align*}
where, 
$$
\mathbb{P}( f(X) 
    < \ell)  = q \; \mathbb{P}(\delta^{+} \le \ell)  + (1-q) \; \mathbb{P}(\delta^{-} \le \ell),
$$
and,
$$
 \mathbb{P}(f(X) > h) = q \; \mathbb{P}(\delta^{+} \ge h) + (1-q) \; \mathbb{P}(\delta^{-} \ge h). 
 $$
Let $\phi^{+}(\cdot)$ and $\phi^{-}(\cdot)$ be the empirical cumulative distributions of $\delta^{+}$ and $\delta^{-}$, respectively. Then, the workload constraint \eqref{cc:4} can be written as 
\begin{align} \label{c:workloadCDF}
    q \left( \phi^+(h) - \phi^+(\ell) \right) + (1-q) \left( \phi^-(h) - \phi^-(\ell) \right) \; \le \; \rho.
\end{align}

We note that increasing $\ell$ weakly decreases the left-hand side of the workload constraint while weakly increasing the first term in the objective function. Similarly, decreasing $h$ weakly decreases the left-hand side of the workload constraint while weakly increasing the second term in the objective function. Thus, when $h(\xi^{ru}) > \ell(\xi^{as})$ and $q \left( \phi^+(h(\xi^{ru})) - \phi^+(\ell(\xi^{as})) \right) + (1-q) \left( \phi^-(h(\xi^{ru})) - \phi^-(\ell(\xi^{as})) \right) \; > \; \rho$, the optimal thresholds should ensure that the workload constraint holds with equality. Under those conditions, the \textcolor{black}{Primary Problem} in Section~\ref{policy description} can be reformulated as follows:
\begin{align*} 
   \min_{\ell,h}   &\;\; \lambda \; \phi^{+}(\ell) + (1-\lambda) \; \left(1-\phi^-(h) \right)\\
    \text{s.t.}&\;\; q \left( \phi^+(h) - \phi^+(\ell) \right) + (1-q) \left( \phi^-(h) - \phi^-(\ell) \right) = \rho \\
    &\;\; \ell(\xi^{as}) \; \le \; \ell \; \le \; h \; \le \; h(\xi^{ru}).
\end{align*}

For the above optimization, it is still not possible to compute a closed-form solution as the workload constraint is coupled with the objective. However, we can use a nested search approach to identify the optimal thresholds \(\ell\) and \(h\). Specifically, we treat \(\ell\) as the primary decision variable. For each candidate \(\ell\), we then solve for an $h^*(\ell)$ that enforces the workload constraint in equality (provided such \(h\) exists). The procedure proceeds in two key steps:

\medskip
\noindent\textbf{Step 1 (Inner sub-problem).} 
For a given \(\ell\in\bigl[\ell(\xi^{as}),\,h(\xi^{ru})\bigr]\), we seek \(h^*(\ell)\in[\ell,\,h(\xi^{ru})]\) such that
\begin{align}
  q\,\bigl(\phi^+\bigl(h^{*}(\ell)\bigr) - \phi^+(\ell)\bigr)
  \;+\;
  \bigl(1-q\bigr)\,\bigl(\phi^-\bigl(h^{*}(\ell)\bigr) - \phi^-(\ell)\bigr)
  \;=\;
  \rho \, ,
  \label{eq:nested_workload}
\end{align}
where for some large values of \(\ell\), the above equation might be infeasible. To find such a solution, we employ a bisection search as follows:

\begin{enumerate}
  \item Initialize \(\bigl[h_L,h_R\bigr]=\bigl[\ell,h(\xi^{ru})\bigr]\).
  \item Define 
  \[
    g(\ell,h) \;=\; q\,\bigl(\phi^+(h)-\phi^+(\ell)\bigr) \;+\;\bigl(1-q\bigr)\,\bigl(\phi^-(h)-\phi^-(\ell)\bigr) - \rho \, .
  \]
  \item While \(\,|h_R - h_L|\) is above a tolerance $\zeta$, evaluate 
  \(\,h_{\mathrm{mid}}=\tfrac{1}{2}(h_L + h_R)\,\) and then:
  \[
    \begin{cases}
       h_L = h_{\mathrm{mid}}, & \text{if } g\bigl(\ell, h_{\mathrm{mid}}\bigr) < 0 \, ,\\[3pt]
       h_R = h_{\mathrm{mid}}, & \text{otherwise}.
    \end{cases}
  \]
  \item The final solution, $h^{*}_{\zeta}(\ell)$, is the midpoint of \(\bigl[h_L,h_R\bigr]\).
\end{enumerate}

\medskip
\noindent\textbf{Step 2 (Outer sub-problem).}
We next identify the optimal \(\ell\) within the interval \(\bigl[\ell(\xi^{as}),\,h(\xi^{ru})\bigr]\). We discretize \(\ell\) over a uniform grid \textcolor{black}{with grid step $\Delta$}:
$$\mathcal{L} 
   \;=\; 
   \{\ell_1,\,\ell_2,\dots,\ell_m\} \;\subseteq\; [\ell(\xi^{as}),\,h(\xi^{ru})),
   \quad
   \text{with } m \;=\; \left \lceil \frac{\,h(\xi^{ru}) - \ell(\xi^{as})\,}{\Delta} \right \rceil. $$

We proceed as follows for each $\ell \in \mathcal{L}$:
\begin{enumerate}
    \item Apply the bisection method to compute $h^{*}_{\zeta}(\ell)$. If such a solution exists, add $\ell$ to the set of feasible candidates $\mathcal{L}_{\text{feas}}$; otherwise, discard $\ell$.
  \item Evaluate the objective and record it:
  \[
     F\bigl(\ell\bigr) \;=\; \lambda \; \phi^{+}(\ell) + (1-\lambda) \; \left(1-\phi^-(h^{*}_{\zeta}(\ell)) \right).
  \]
\end{enumerate}
Finally, select the thresholds as
 $$ \ell^{NS}\;=\;\arg\min_{\ell \in \mathcal{L}_{\text{feas}}} F\big( \ell \big),
  \quad
  h^{NS} \;=\; h^{*}_{\zeta}\big( \ell^{NS} \big),$$

\noindent and return \(\bigl(\ell^{NS},h^{NS}\bigr)\) as the solution. The overall procedure is summarized in Algorithm~\ref{alg:nested_search} \textcolor{black}{(see Appendix \ref{EC:nested search})}.

Utilizing the structural properties outlined in Section~\ref{Structural Properties} and our nested search approach, we present our main theoretical result, which introduces a systematic approach to solving the \textcolor{black}{Primary Problem}.

\begin{theorem} \label{thm2} For the \textcolor{black}{Primary Problem}, we have: 
\begin{enumerate} [label=(\alph*)]
\renewcommand\labelenumi{\normalfont(\alph{enumi})}
    \item If $h(\xi^{ru}) \;>\; \ell(\xi^{as})$ and  the workload constraint \eqref{c:workloadCDF} holds for $\ell \,=\,  \ell(\xi^{as})$ and $h \,=\, h(\xi^{ru})$, then $(\ell(\xi^{as}), h(\xi^{ru}))$ is an optimal solution,
    \item If $h(\xi^{ru}) \;>\; \ell(\xi^{as})$ and the workload constraint \eqref{c:workloadCDF} does not hold for $\ell \,=\,  \ell(\xi^{as})$ and $h \,=\, h(\xi^{ru})$, then for any given $\epsilon > 0$, \textcolor{black}{Algorithm \ref{alg:nested_search} (Nested Search)} with \textcolor{black}{grid step $\Delta$ and tolerance $\zeta$} returns an $\epsilon$-optimal solution after at most 
    $$ \mathcal{O} \left( \frac{ \bar{L} \left(  h(\xi^{ru}) - \ell(\xi^{as}) \right)}{\epsilon}  \; \log_2 \left( \frac{ 2 \bar{L}_{\phi} (1-\lambda) \left(h(\xi^{ru})-\ell(\xi^{as}) \right)}{\epsilon}  \right) \right)$$ 
    iterations, where $\lambda \in (0,1)$, $\bar{L}_{\phi}$ and $\bar{L}_{h}$ are upper bounds on the common Lipschitz constant of $\phi^{\pm}(\cdot)$ and the Lipschitz constant of $h^{*}(\ell)$ respectively, and  $\bar{L} = \bar{L}_{\phi} \left(\lambda + (1-\lambda)\bar{L}_h \right)$. Setting $\Delta \; = \; \epsilon/[2\bar{L}]$ and $\zeta \;=\; \epsilon/[2(1-\lambda)\bar{L}_{\phi}] $ guarantees $\epsilon$-optimality and ensures that the workload constraint is satisfied within a tolerance of $\epsilon/[2 (1-\lambda)]$,
    \item If $h(\xi^{ru}) \;<\; \ell(\xi^{as})$, then the problem is infeasible,  
    \item If $h(\xi^{ru}) \;=\; \ell(\xi^{as})$, then the problem has a single threshold solution $\ell^* \;=\; h^* \;=\; h(\xi^{ru}) \;=\; \ell(\xi^{as}).$
\end{enumerate}
\end{theorem}

\color{black}
\section{Policy Recommendation and Impact Evaluation} \label{Empirical Results}
Using our results from the previous section, we now propose a data-driven clearance policy that leverages a human-algorithm approach to assist the FDA in its decision-making. We also leverage the dataset we have assembled (see Section~\ref{Setting and Data}) and investigate the effectiveness of our policy vis-a-vis the FDA's current practice. \textcolor{black}{To this end, we utilize the same stratified random split employed in Section \ref{Prediction}, dividing the data into training (70\%) and testing (30\%) sets.} We use the ML model presented in Section \ref{predictive model results} to provide the inputs required for the optimization procedures discussed in the previous section.

When making use of our policy, three parameters should be set by the decision-maker: $\xi^{ru}$, $\xi^{as}$, and $\rho$, which correspond to the rates of rejection of unsafe devices, the acceptance of safe devices, and the upper bound on the workload, respectively. 
This can assist decision-makers in selecting the right input parameters that align with their criteria. We investigate the impact of these parameters on the acceptance and rejection rates of both safe and unsafe devices as well as the FDA's workload, which currently stands at 100\% as all devices are evaluated by the FDA's committees. To provide a clear performance evaluation, we consider a conservative scenario where we assume that the \textcolor{black}{additional information} for deferred devices do not contribute to enhancing the evaluation process conducted by FDA committees. \textcolor{black}{For any device that our policy defers to the FDA committee, we assume the committee takes the same action it took historically, and we evaluate safety using that device’s observed recall outcome.} Thus, our reported estimates of potential impact are conservative since it is likely that the \textcolor{black}{additional information for deferred devices} can themselves enable the FDA's human experts to improve their decisions. To do so, each input parameter is considered at three levels: low (L), medium (M), and high (H). For the parameters $\xi^{ru}$ and $\xi^{as}$, we have $L = 0.3$, $M = 0.5$, and $H = 0.7$. For the parameter $\rho$, the values are $L = 0.4$, $M = 0.6$, and $H = 0.8$. A summary of results is presented in Appendix \ref{ImpactofInput}. The results indicate that there is an interplay between the threshold parameters and their influence on the acceptance and rejection rates for safe and unsafe devices. A higher value of $\xi^{ru}$ leads to a more stringent policy, resulting in an increased rate of rejection for safe devices. Similarly, a higher value of $\xi^{as}$ is associated with a corresponding increase in the acceptance rate of unsafe devices. This highlights the intricate balance required between safety and efficiency in the policy. 

\begin{figure} [tbh] 
    \centering
    \includegraphics[width=0.65\linewidth]{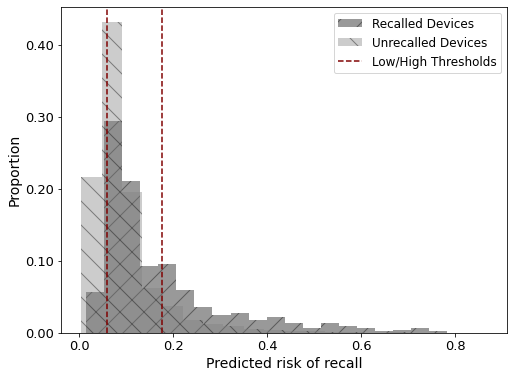}
    \caption{Optimized low and high thresholds corresponding to $\xi^{ru} = L$, $\xi^{as} = L$, and $p = M$. Distributions of predicted risk values for the unrecalled devices and recalled devices are shown with right- and left-hatched patterns, respectively.} 
    \label{fig8}
\end{figure}

\textcolor{black}{To gain deeper insights, we next focus on the L-L-M combination as a representative setting. This combination corresponds to setting the minimum performance parameters ($\xi^{ru}$ and $\xi^{as}$) to Low (0.3), demonstrating that even with minimal mandatory performance requirements, our policy has the potential to significantly outperform the FDA's current practice. Additionally, this combination sets the maximum workload parameter ($\rho$) to Medium (0.6). This target strikes a  balance between efficiency (achieving up to a 40\% reduction in workload) and the continued necessity of human oversight, avoiding aggressively forcing too many automated decisions. Ultimately, however, the specific combination of parameters would need to be determined by the FDA based on their operational preferences and resource constraints.} Figure~\ref{fig8} illustrates the optimized low and high thresholds, as well as the distribution of predicted risk values for unrecalled and recalled devices. The right-hatched region (\textbackslash) to the left of the low threshold represents the proportion of safe devices correctly accepted by the policy, while the right-hatched region (\textbackslash) to the right of the high threshold represents the proportion of safe devices falsely rejected. Similarly, the left-hatched region (/) to the right side of the high threshold represents the proportion of unsafe devices correctly rejected by the policy, while the left-hatched region (/) to the left of the low threshold represents the proportion of unsafe devices falsely accepted. Additionally, the overlapping hatched region between the low and high thresholds shows the proportion of deferred devices.

Table~\ref{table44} shows the impact of the representative policy in the testing set. As can be seen, using this policy leads to a 40.5\% reduction in the workload of the FDA's committees and a 32.9\% improvement in the recall rate percentage. Investigating the policy, we observe that applicant devices with a recall risk estimated to be below 0.060 are accepted, while applicant devices with a recall risk estimated to be higher than 0.177 are rejected. The other applicant devices are deferred for judgment by an FDA committee of human experts. Under our conservative evaluation, following this policy results in the rejection of 32.9\% of unsafe devices and the acceptance of 30.7\% of safe devices. Among the accepted devices, 8.9\% will experience a recall in the future, while 9.7\% of the rejected devices will have no recalls in the future. \textcolor{black}{The overall rejection rate of the policy is 12.1\%. When decomposed relative to the total applicant pool, this consists of 3.4\% correct rejections and 8.7\% false rejections. While the overall rate is higher than the historical 5\% NSE rate (\citealt{CDRH}), this increase is a justifiable trade-off given the 32.9\% improvement in the recall rate. This significant enhancement directly prevents patient harm and contributes to substantial annual cost reductions, as detailed in Section~\ref{Policy Impact: Costs}. Furthermore, our policy offers a range of solutions on the efficient frontier; if FDA deems this rejection rate too high, the policy parameters can be adjusted to reduce rejections, though this would result in a lower improvement in the recall rate.}

\begin{table} [tb]
\centering \small
\caption{Impact of the representative policy on workload reduction and improvement in recall rate percentage compared to the FDA's current practice} \label{table44}
\begin{tabular}{lllll}
\hline
\textbf{$ \ell $} & \textbf{$ h $} & \textbf{\begin{tabular}[c]{@{}l@{}}Reject~\\Rate (\%)\end{tabular}} & \textbf{\begin{tabular}[c]{@{}l@{}}Workload~\\Reduction (\%)\end{tabular}} & \textbf{\begin{tabular}[c]{@{}l@{}}Recall Rate~\\Improvement (\%)\end{tabular}} \\
\hline
0.060 & 0.177 & 12.1 & 40.5 & 32.9 \\
\hline
\end{tabular}
\end{table}

A robustness check is also conducted to assess the potential impact of selection bias that may arise from the absence of information on devices rejected by the FDA. Our findings suggest that including such devices does not undermine the overall effectiveness of our proposed policy (see Appendix \ref{EC:SelectionBias} for details). \textcolor{black}{Additionally, we conduct a post-hoc analysis comparing the characteristics of devices deferred, accepted, or rejected by our approach, and contrasted devices accepted by our policy with those accepted under current practice. Our findings confirm that deferred devices typically exhibit borderline risk signals warranting human review, while devices accepted by our policy generally demonstrate safer predicate histories compared to current practice. Detailed results and discussions of this analysis are provided in Appendix~\ref{Post-Hoc Analyses}.}

\subsection{Benchmark Comparison: ML-Only vs. Human-Algorithm Policies} \label{sec:benchmark_comparison}
In addition to evaluating our policy against the FDA's current practice, it can be insightful to compare it to a standard, single-threshold ML-only policy. 

First, our conservative analysis under a fixed set of preferences highlights a critical operational trade-off (see Appendix \ref{EC:Comparison} for details). The ML-Only policy (corresponding to $\rho=0$) achieves high recall but with an impractically high overall rejection rate. Our policy, by varying $\rho$, provides a crucial operational lever. This creates a curve of operationally feasible alternatives, allowing a decision-maker to trade a portion of that recall to dramatically reduce the rejection rate and the FDA's workload.

Second, a more robust comparison of the policies' entire performance frontiers can be conducted by varying the priority-weighting $\lambda$. However, this comparison presents challenges. A conservative setting, which assumes the FDA's performance in evaluating deferred cases will not improve under our policy, does not fully capture the potential value of our human-algorithm approach. In particular, because our dataset contains only devices that were ultimately cleared (accepted) by the FDA, our conservative evaluation assumes that all deferred devices will be accepted as well. Consequently, for any of our policy's thresholds $(\ell, h)$, an ML-only policy can simply set its single threshold $t = h$. This makes the two policies operationally identical, as they reject the same devices (risk $>h$) and accept all others.

In practice, unlike an ML-only policy, our policy defers a proportion of devices to the FDA committee. Accordingly, the performance of our policy depends partly on the performance of that committee, which could be improved in two main ways: (i) the committee receives additional information for deferred devices, aiding in their evaluation, and (ii) direct recommendations for non-deferred cases can potentially reduce the committee's workload, allowing more resources for the in-depth evaluation of these complex cases. To provide a meaningful benchmark, Appendix \ref{EC:Comparison} presents a non-conservative analysis in which we model the added value of the deferral process. This approach relies on the assumption that the additional information and potential workload reduction provided by our policy would enable the FDA committee to better detect unsafe devices. The results of this detailed comparison show that, under these assumptions, our human-algorithm policy is strictly dominant, consistently and substantially outperforming the best-possible ML-only frontier. This demonstrates that identifying complex cases with our ML model and deferring them to human experts leads to a superior policy overall.

\color{black}
\subsection{Policy Impact: Costs} \label{Policy Impact: Costs}
In the previous sections, we evaluated the impact of our proposed policy in terms of measures such as correct acceptance and rejection rates as well as the resulting workload. In this section, we examine its impact on costs. To this end, we note that a number of independent studies have shown that recalled or prematurely failed devices have likely cost Medicare billions of dollars in recall-related health care expenditures (\citealt{USdephealth2017}). In one year alone, the FDA received reports of nearly three-thousand potential device-related deaths, over one-hundred thousand potential device-related injuries, and over two-hundred thousand adverse event reports concerning medical devices (\citealt{zuckerman2011medical}). 
Nonetheless, the full extent of injury to patients that is attributable to recalled devices is unknown. \textcolor{black}{Our focus is on overall cost-savings of our proposed policy. Thus, we conservatively assess the cost savings based on the replacement costs (approximated by Medicare allowed amounts, which include Medicare payments and patient deductibles) of a recalled device.}

It is worth noting that the true cost of a recall may far exceed our calculations (i.e., the replacement costs incurred by the manufacturers) because it will also include the potential cost of related injuries. In 2016, for example, a settlement of various state and federal personal injury litigations from recalled pelvic mesh products totaled \$121 million (\citealt{Medtronicplc2023}). In a separate case in 2016, personal injury claims concerning a bone graft product were settled for \$26 million (\citealt{Medtronicplc2016}). The fact that these amounts reflect settlements, not affirmative decisions by courts, suggests that full compensation for personal injury could have been even higher. McKinsey estimated that non-routine events ``such as major observations, recalls, warning letters, and consent decrees, along with associated warranties and lawsuits” cost the industry between \$2 billion and \$5 billion per year on average. The total cost includes \$1.5 billion to \$3 billion per year on non-routine costs, plus \$1 billion to \$2 billion in lost sales of new and existing products. This suggests that annual non-routine costs of recalls can range between \$0.5 billion to \$3 billion per year (\citealt{fuhr2013business}). \textcolor{black}{While our policy reduces recall-related expenditures, it also results in some false rejections related to cases in which a device with no observed recall in our dataset is recommended for rejection. We argue that the cost of such false rejections for 510(k) devices is clinically minimal. Because 510(k) devices must be substantially equivalent to an already-marketed device, most do not provide one-of-a-kind treatments. Thus, if our policy erroneously rejects certain 510(k) devices, near-identical substitutes (i.e., the predicate or a similar cleared device) presumably remain on the market. While the clinical harm to patients is minimal, false rejections impose financial costs on manufacturers (e.g., additional delays or resubmissions) and can have broader economic consequences for payers and other stakeholders that are outside the scope of our analysis.}

Our assessment of the replacement costs for recalled devices is based on administrative claims data titled “Medicare Durable Medical Equipment, Devices \& Supplies” (MDMEDS) for the years 2013-2020. This data is published annually by the Centers for Medicare and Medicaid Services (CMS) and contains information on usage, payments, and submitted charges organized by National Provider Identifier (NPI) and Healthcare Common Procedure Coding System (HCPCS) code (\citealt{CMS1}). The dataset is based on information gathered from CMS administrative claims data for Original Medicare Part B beneficiaries available from the CMS Chronic Conditions Data Warehouse. The data are summarized from 100\% final-action Durable Medical Equipment, Prosthetic, Orthotics and Supplies (DMEPOS) non-institutional claim line items (\citealt{CMS1}). We calculate the replacement costs by medical specialty using the HCPCS codes and descriptions reported in the MDMEDS data. \textcolor{black}{The details of this calculation, which require creating a crosswalk between keywords and medical specialties, are discussed in Appendix \ref{crosswalk}.}

\textcolor{black}{To calculate the potential cost savings, we follow the conservative scenario assuming that the additional information for deferred devices does not contribute to enhancing the evaluation process conducted by FDA committees.} In particular, we calculated the potential cost savings from implementing our data-driven policy derived from the L-L-M combination (see Figure~\ref{fig8}) on the testing set with 9,572 510(k) submissions. In Step 1, we identified 325 recalled devices that were cleared by the FDA but our policy would have rejected if implemented. In Step 2, we determined the number of units being recalled for each of the aforementioned devices using the FDA recall data, which amounted to approximately 47.1 million units. In Step 3, we used the low and high average Medicare allowed amounts to determine the low and high estimated cost savings that would have occurred had these devices been rejected pursuant to our proposed policy.

\begin{figure}[tb]
\centerline{\includegraphics[scale=.5]{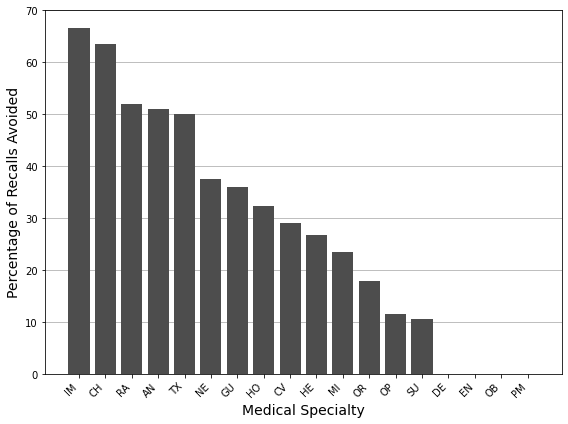}}
\caption{Impact of our proposed policy in terms of avoiding recalls in the testing set}
\label{fig11}
\end{figure}

Figure~\ref{fig11} shows the percentage of recalls avoided by our proposed policy per medical specialty in the testing set (Step 1). For example, 29.1\% for the CV medical specialty indicates that 44 recalls out of 151 recalls corresponding to the CV medical specialty in the testing set were avoided by our proposed policy. As can be seen, Immunology (IM) devices have the highest frequency, followed by Clinical Chemistry (CH) and Radiology (RA). On the other spectrum, Dental (DE), Ear, Nose, \& Throat (EN), Obstetrics/Gynecology (OB), and Physical Medicine (PM) have the lowest frequency (zero). Comparing with the recall rate per medical specialty in our dataset, we observe that EN, OB, and PM are among the top four medical specialties with low recall rates (less than 1\%) and DE has a recall rate of 1.9\%. The alignment between low avoided recalls and low overall recall rates suggests that our policy is functioning as intended, particularly in areas where it is most needed. Table~\ref{table9} reports the total average cost savings associated with the four most cost-saving medical specialties (Step 3). Among them, Orthopedic (OR) is the medical specialty with the highest average cost savings.  

The results based on our testing set show that the overall cost savings from implementing our policy range between \$5.5 billion and \$5.6 billion over the several years of our study period. Given that there are approximately 3,000 510(k) submissions annually (\citealt{dubin2021risk} and \citealt{kadakia2023use}), by extrapolating from the total cost savings in our testing set, a rough estimate of the annual cost savings is \$1.7 billion. \textcolor{black}{To calibrate the magnitude of the estimated \$1.7 billion in annual cost savings, we compared it against the total annual U.S. medical device market size, which includes all devices with the vast majority being non-recalled. Based on historical data from the U.S. Department of Commerce's International Trade Administration, the U.S. medical device market was valued at approximately \$156 billion in 2017 (\citealt{ITA2016MedicalDevices}). We utilize this figure as the representative average for the period between 2013 and 2020 to ensure consistency with our cost data (derived from 2013–2020 claims). Calibrating our \$1.7 billion savings against this total market average indicates that our cost savings are equivalent to approximately 1.1\% of the total U.S. medical device expenditure.}

\definecolor{OxfordBlue}{rgb}{0.207,0.247,0.317}
\begin{table}[tb]
\centering \small
\caption{Total cost savings associated with four most cost-saving medical specialties in the testing set}
\label{table9}
\begin{tblr}{
  hline{1-2,6} = {-}{},
}
\textbf{Medical specialty }    & \textbf{Recalls Avoided \%} & {\textbf{Average} \\\textbf{Product Quantity}} & {\textbf{Average cost savings} \\\textbf{(in 1,000 U.S dollars)}}  \\
Orthopedic (OR)       & 17.9               & 179,229.3                    & \$ 2,276,838                                     \\
General Hospital (HO) & 32.3               & 855,095.4                  & \$ 1,076,394 \\
Anesthesiology (AN)   & 51.0               & 603,998.7                    & \$ 990,606 \\
Cardiovascular (CV)   & 29.1               & 8,824.2                     & \$ 959,569             
\end{tblr}
\end{table}

\color{black}
\section{\textcolor{black}{Managerial Insights and Conclusion}}
\label{ManagerialInsights}
Our work investigates the degree to which a data-driven policy can assist the FDA in improving its 510(k) medical device clearance process. We hypothesized that our combined human-algorithm approach to evaluating medical devices will result in a reduction of recall rates and the workload burden imposed on the FDA. We conducted an in-depth evaluation of the performance of our policy as well as the FDA's current practice.
The results confirm our hypothesis that there is a need for a combined human-algorithm approach, where devices with a mid-range predicted risk of recall (non-easy cases) are deferred to human experts for further evaluation. That is, integrating the FDA's expertise with quantitative evidence is required to improve the 510(k) medical device clearance process. A conservative evaluation of our proposed policy based on our data showed a 32.9\% improvement in the recall rate and a 40.5\% reduction in the FDA's workload. 
\textcolor{black}{Our cost analysis focuses on savings to the healthcare system from avoided replacement costs, approximated by Medicare allowed amounts (including Medicare payments and patient deductibles). We projected that implementing our policy could result in significant annual cost savings of \$1.7 billion. This amount represents roughly 1.1\% of the total annual U.S. medical device expenditure, which is a significant reduction given the scale of the market.
We note that the cost of potential false rejections resulting from the implementation of our policy is clinically minimal, as the predicate or a similar device presumably remains on the market. Additionally, the lower rate of false rejections assures that the unmodeled financial costs to manufacturers and payers are likely under control. Overall, these findings imply that our proposed policy is effective in significantly reducing recall rates, the FDA's workload, and costs incurred due to recalls.}

We believe that our work addresses some of the main concerns about the current 510(k) process. Most recently, the FDA has responded to these concerns with modernization efforts “to improve the safety of medical devices while continuing to create more efficient pathways to bring critical devices to patients” (\citealt{FDASafetyActionPlan}). In 2018, the FDA proposed “promoting innovation and improving safety by driving innovators toward reliance on more modern predicate devices.” That is, the newer devices should be compared to the benefits and risks of more modern technology as the older predicates might not reflect the advanced technology embedded in new devices or align with the FDA's current understanding of device benefits and risks (\citealt{FDAStatement}). 
However, the FDA ultimately acknowledged that its initial proposal “may not optimally promote safer and more effective devices” in all instances since devices that use modern rapidly-evolving technology could benefit from comparison to more recent predicates whereas older devices may establish a history of safety and effective use in other cases. 
\textcolor{black}{We found that a higher value for the newest predicate age is linked to a reduced recall risk. This is possibly due to the limited real-world usage of newer predicates, which may not have been sufficient to reveal potential issues. Regarding the oldest predicate age, our results indicated that higher values can lead to a lower recall risk. This aligns with the idea that older predicates can indicate safety, as they have a history of safe and effective use and may be the gold standard for patient care. We also found that higher values of the median predicate age correlate with a lower recall risk. Altogether, these findings highlight a learning curve effect. That is, if a predicate device has been on the market only briefly, latent problems may remain undiscovered, leading to a higher likelihood of future recalls.} 

Recently, the FDA developed a new draft guidance in 2023 on best practices for selecting predicate devices based on device characteristics rather than just age (\citealt{FDABestPractices}). Three of the FDA’s proposed best practices include (1) selecting predicates that continue to perform safely and as intended, (2) selecting predicates that do not have unmitigated use-related or design-related safety issues, and (3) selecting predicates that have not been subject to a design-related recall. We found that the number of recall events for predicates is a significant predictor of recall risk for an applicant device. This finding specifically aligns with the FDA's proposed practice regarding the selection of predicates that continue to perform safely. Additionally, our results emphasize the importance of the timing of predicates' recall events. These insights call for a more targeted approach towards predicates with recent recalls. Our hypothesis is that manufacturers might not have had sufficient time to address potential issues with these predicates. Consequently, applicant devices resembling such predicates may face recall due to similar issues.
Shifting focus to the characteristics of the applicant devices, we observed that some product codes, country codes, and medical specialties are crucial predictors of recall risk. The results are consistent with prior literature suggesting the heterogeneous effects of these indicators on the risk of recall. Furthermore, our study confirms the importance of variables that quantify risk, such as life-sustaining.
Our conjecture is that riskier applicant devices are generally under stringent market scrutiny and are more likely to be recalled. 


\textcolor{black}{Our proposed policy has four specific benefits. First, it has}
a transparent structure (two phases based on clear rules), which allows for continued transparency in the 510(k) review process. Second, it benefits from explainability, an important element in allowing
stakeholders, such as regulators and health providers, to sanity check models beyond mere performance (\citealt{amann2020explainability}). 
Third, in developing and evaluating our policy, we paid specific attention to selecting appropriate metrics, such as false positive and false negative rates and the FDA's workload. 
We highlight that relying solely on one of these metrics may not provide a complete picture. For example, false positives and false negatives can have significantly different consequences. False positives may lead to the unnecessary rejection of safe devices, while false negatives could result in missed diagnoses of unsafe devices. 
\textcolor{black}{Finally, our proposed policy builds upon the established framework of human-in-the-loop decision-making, where deferring uncertain decisions to human experts can improve overall system performance. This approach allows us to leverage the complementary strengths of AI algorithms and human judgment, a synergy that recent healthcare literature has highlighted as an effective approach for clinical and regulatory tasks (\citealt{orfanoudaki2022algorithm}, \citealt{saghafian2023effective}). }
\textcolor{black}{At the same time, the policy preserves regulatory safety by supporting the FDA’s substantive review of devices that have already passed initial administrative and eligibility checks, such as verification of intended use and technological characteristics.} To keep the human-algorithm partnership improving, one useful approach is to establish an oversight committee, separate from the review committees, that meets periodically to audit recent approvals, rejections, and deferrals, hear stakeholder feedback, and examine extreme cases flagged by our policy. The committee can then adjust the decision thresholds as needed, allowing the system to improve continuously over time.


\textcolor{black}{It is worth discussing possible concerns regarding the practicality of implementing our proposed policy, including (i) FDA’s ability to leverage our data-driven clearance policy, (ii) transparency versus \textit{gaming}, and (iii) heterogeneity across medical specialties.} Regarding point (i), we note that the FDA’s recent draft guidance on best practices for selecting a predicate device indicates that our policy would align with the FDA’s avowed commitment “to improve the predictability, consistency and transparency” of the 510(k) pathway while not proposing changes to applicable statutory and regulatory standards, such as how the FDA evaluates substantial equivalence, or the applicable requirements, including the requirement for valid scientific evidence” (\citealt{FDASafetyActionPlan}). As discussed earlier, three of the FDA’s proposed best practices are an attempt to improve the safety of medical devices by mitigating the use of predicates with safety issues. Our data-driven policy uses established AI/ML methods to provide additional scientific evidence, namely an estimated recall risk of a 510(k) applicant device based, in part, on safety issues present in predicate devices. Moreover, our clearance policy is not intended to be a substitute for the FDA’s expertise in assessing applicant devices. It is a tool that the FDA can leverage, as it sees fit, to increase device safety while reducing workload. The FDA, among other approaches, can choose to use our tool as a guide for identifying devices that warrant higher priority concerns rather than outright rejection. 

Regarding point (ii),
\textcolor{black}{we note that some level of transparency is desirable as it steers manufacturers toward inherently safer designs and decreases the review time. However, complete disclosure of every model feature or weight may invite gaming (\citealt{wang2023algorithmic}). For example, applicants might begin to mask factors that increase the risk of rejection. A practical compromise is to publish the core risk factors while keeping the finer details of the prediction model and the risk thresholds of the policy proprietary to the FDA. The policy itself will be adapted dynamically, which allows detecting and counteracting some masking tactics. It is worth noting that our estimates of reductions in recall events and FDA's workload rest on the empirical distribution of 510(k) submissions observed in our dataset. If our proposed policy alters manufacturers’ strategic behavior, for example by deterring some higher-risk devices or attracting additional low-risk devices, the post-implementation submission mix could differ from our estimation sample. This could introduce endogeneity and bias the projected gains. Periodic recalibration of the policy may help the algorithm adapt to emerging patterns and mitigate drift arising from manufacturers’ adaptive behavior, but it may not fully eliminate such effects.}
\textcolor{black}{Lastly, regarding point (iii), we note that our current policy applies a uniform set of risk thresholds ($\ell$ and $h$) across all devices to prioritize structural simplicity. Nonetheless, the performance of our predictive model and the underlying risk profiles differ by medical specialty. Consequently, the policy’s operational impact, such as the specific rate of recalls avoided, may vary between specialties. However, our methodological framework is flexible enough to accommodate specialty-specific considerations. In practice, there may be value in refining the policy by tuning the upper and lower risk thresholds as a function of medical specialty or by ensuring risk scores are calibrated at the specialty level.}

\textcolor{black}{We believe future studies are needed to further investigate the impacts of implementing our policy, such as examining potential changes to the predicate selection process or conducting randomized experiments to validate performance. Additionally, future research could explore the use of AI/ML tools to enhance the FDA's initial eligibility checks. Finally, beyond the immediate FDA context, our human-algorithm framework holds promise for other domains that must sift through large applicant pools under tight resource constraints, such as university admissions, where data-driven screens could formalize the review of clear-cut cases while reserving human expertise for borderline files.}

\color{black}

\ACKNOWLEDGMENT{The authors would like to thank Anders Olsen for his assistance with data curation. They also thank George Ball for his thoughtful feedback on the manuscript. Additionally, the authors are grateful for the valuable comments and helpful suggestions received during the presentation of this work at several conferences. Finally, the authors would like to thank the Department Editor, Professor Stefan Scholtes, the anonymous Associate Editor, and the anonymous referees for their constructive and detailed comments, which have significantly improved both the content and the exposition of this paper.
}

\def\bibfont{\scriptsize}
\bibliographystyle{informs2014}
\bibliography{references}

\ECSwitch

\begin{center}
{\large \bf Electronic Companion to }\\
{
{\large
{\bf
``Harmonizing Safety and Speed: A Human-Algorithm Approach to Enhance the FDA’s Medical Device Clearance Policy"
}
}}
\\
\vskip 0.05in
{\normalsize
by Mohammad Zhalechian, Soroush Saghafian, and Omar Robles}
\end{center}

\bigskip
\noindent This electronic companion comprises all the proofs of our theoretical findings, along with supplementary details and additional results from computational experiments.
\bigskip
\begin{APPENDICES}

\section{Text Mining Algorithm} \label{EC:TextMining}
In this section, we describe the process of mining the predicate data for devices cleared through the 510(k) pathway. It consists of five main steps, each of which is described in detail below.

\noindent \textbf{Step 1: Downloading the 510(k) files of cleared devices.} The FDA maintains downloadable files containing information about devices cleared under the 510(k) process from 1976 to the present. These files are updated monthly, typically on or around the 5th of each month. 
Each record includes the submission identifier (the 510(k) identifier), along with other applicant and device information such as the applicant's name, FDA decision date, and product code. However, predicate information is not included. These files are publicly available without restriction on the FDA website. We merged all available files to create a comprehensive list of 510(k) devices cleared between 1976 and 2020, storing the information in a CSV file.

\noindent \textbf{Step 2: Downloading the summary document for each cleared device.} Predicate information for each cleared device is typically available in a summary document housed within the FDA data warehouse. Unlike the 510(k) files of cleared devices, these summary documents are not available for bulk download. Instead, each device has a webpage that contains a link to the summary document for that device.

The device webpage is accessible through an FDA searchable database. Figure~\ref{figEC-510KPage} shows an example search for device K192656 using the user interface.

\begin{figure}[b]
\centerline{\includegraphics[scale=.3]{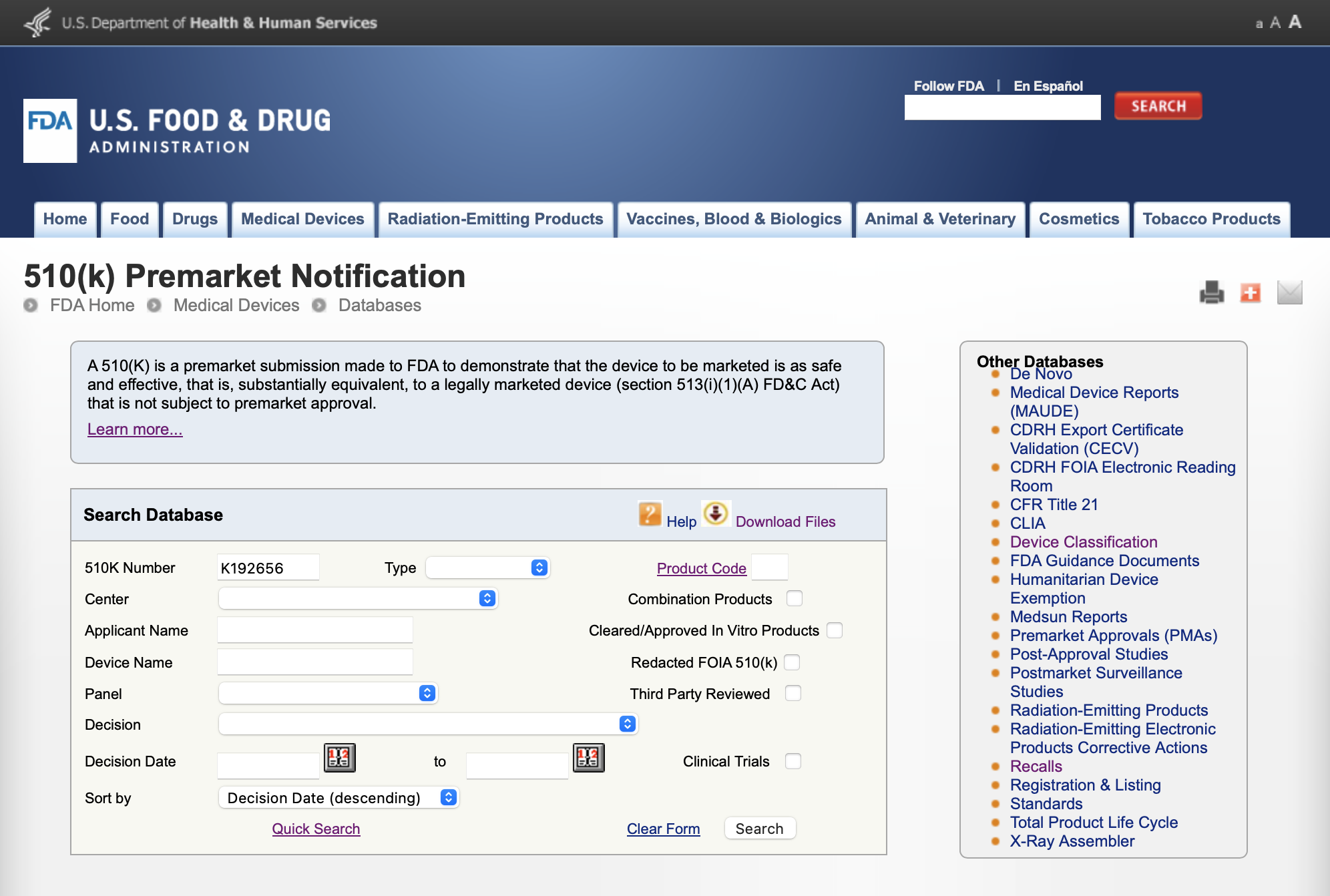}}
\caption{FDA 510(k) Device Search Interface}
\label{figEC-510KPage}
\end{figure}

The search results, as shown in Figure~\ref{figEC-DevicePage}, contain much of the same information available in the downloadable files from Step 1, with one notable exception: the webpage includes a link labeled “Summary,” which typically contains predicate information for that device. Each device's webpage follows a similar URL structure, differing only in the unique 510(k) identifier assigned to each submission.

\begin{figure}[tb]
\centerline{\includegraphics[scale=.3]{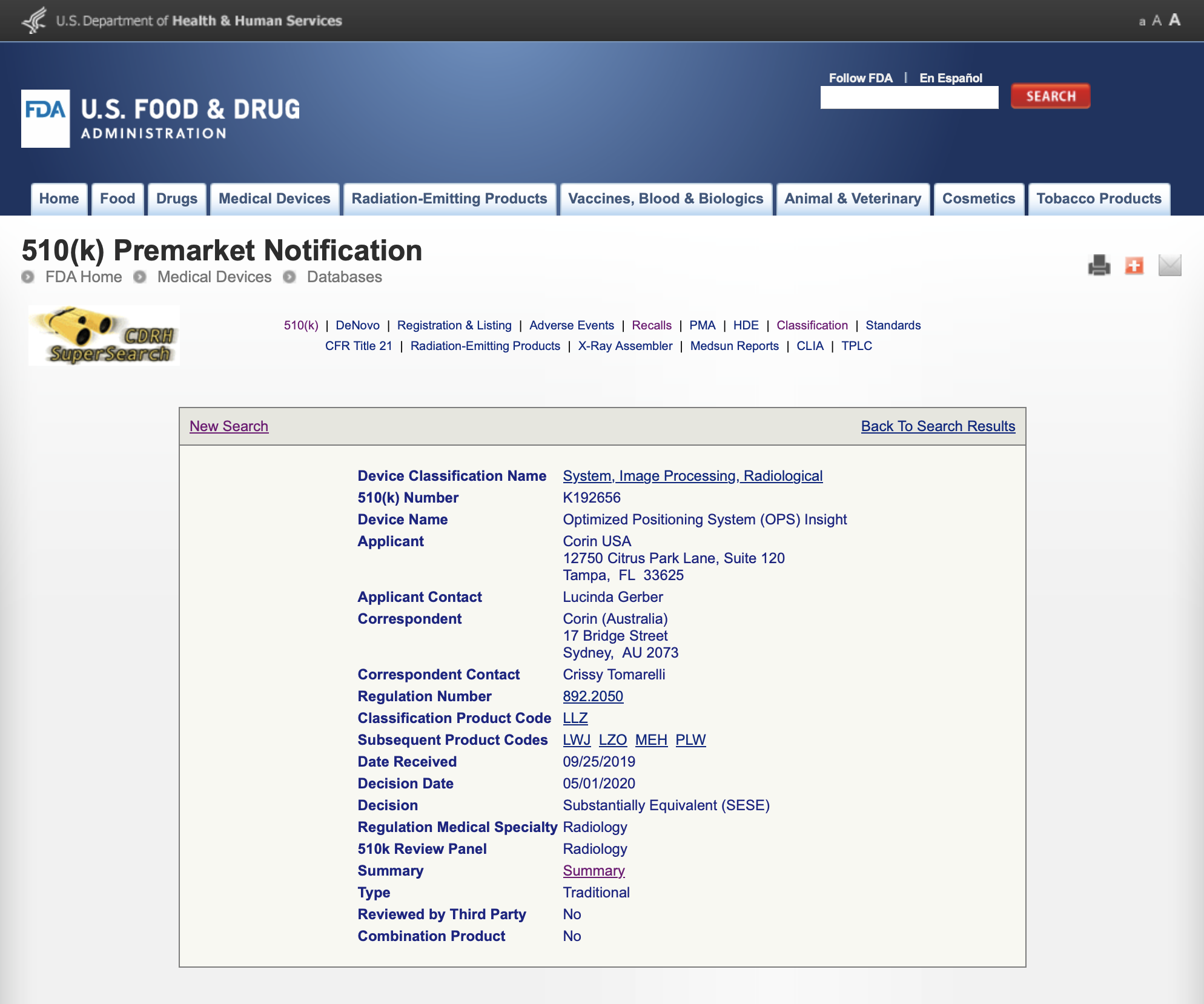}}
\caption{Search results for device K192656}
\label{figEC-DevicePage}
\end{figure}

To automate the retrieval process and eliminate the need for labor-intensive manual searches, we developed a Python algorithm that uses the CSV file generated in Step 1 to download each device’s respective summary document. The algorithm uses the ``requests", ``BeautifulSoup", and ``urllib.request" libraries to iteratively access the URL, search for a link containing the term “Summary,” and download the embedded document. The document is then saved as a PDF, and the CSV file is updated to indicate whether a document was successfully downloaded using the 510(k) identifier as a reference. 

\noindent \textbf{Step 3: Scraping the predicate identifier from each summary document.} The summary documents vary in format and quality. Some of them are not readily searchable using optical character recognition (OCR), some lack predicate 510(k) identifiers but provide predicate names, and some contain no predicate information. To address these challenges, we developed a mechanism within our algorithm to handle certain issues automatically, while others are resolved manually.

In general, the algorithm iteratively searches for predicate 510(k) identifiers within a summary document and records them in a CSV file. It utilizes the following main libraries: ``pdfplumber", ``Image", ``pyocr". The algorithm processes the PDF documents downloaded in Step 1. For each PDF document, it searches for a predicate identifier, which follows a standard format consisting of the letter ``K" or ``P" followed by six digits (e.g., K032245, P230042). Devices cleared through the 510(k) pathway have identifiers that start with the letter ``K", while those cleared through the premarket approval (PMA) pathway have identifiers that start with the letter ``P".

Once an identifier is found, the algorithm records it in the CSV file associated with the device's summary document. If the algorithm fails to extract at least one predicate identifier from a summary document, it converts the document into an image for OCR processing before making a second attempt to locate and record the predicate identifier(s). If the algorithm is unable to retrieve a predicate identifier from the summary document, it records a blank space in the CSV file.

\noindent \textbf{Step 4: Manually updating missing predicate identifiers.} In cases where the algorithm fails to record at least one predicate identifier, we manually review the summary document and record the predicate identifier(s) if available. Occasionally, the summary document may list the descriptive name(s) of the predicate device(s) but not the predicate identifier(s). In a smaller fraction of instances, the predicate identifier(s) may contain typographical errors, such as a missing digit or an extra space (e.g., K12345 instead of K12 3456). In either case, we search for predicate device information manually.

To ensure confidence in the identification of predicate devices, we apply the following criteria: (1) the name (or identifier) of the predicate device must match the name (or identifier, excluding any typographical errors) in the summary document, and (2) the predicate device must have been cleared prior to the device for which it serves as a predicate. This approach prevents the misidentification of predicate devices that are different (e.g., newer models within a product line).

Although predicate information is a requirement in the FDA 510(k) application, it is not always included in the summary document. In these cases, the summary document only includes a general statement that the device is substantially equivalent to a previously cleared or approved device without specifying the exact predicate device.
Fortunately, in the vast majority of cases, the summary document contains predicate information, which allowed us to identify the relevant device. 

\noindent \textbf{Step 5: Manually updating predicate identifiers when more than 10 predicates were recorded.} In cases where the algorithm recorded more than 10  predicates, we manually review the summary document. Although no errors were found in this process, we would update the record of predicate identifier(s) had an error occurred. 

\textcolor{black}{While we conducted spot checks to ensure the correctness of identified predicates, we acknowledge that there is a chance that some additional predicates may have gone undetected in a small number of cases.}

\section{Variable Definitions} \label{EC:VariableDefinitions}
 The FDA classifies each device into several medical specialties, such as Orthopedic (OR) or Cardiovascular (CV). \textit{Medical Specialty} refers to the medical specialty of the applicant device identified by the FDA.  
 The FDA classifies medical devices into three classes based on their risks and regulatory controls. Class I devices pose the lowest risk to patients, while Class III devices pose the highest risk. \textit{Device Class} indicates the FDA device class. Most devices in the 510(k) program fall into Class I or Class II. A limited number of applicant devices (preamendments devices) were initially regulated as Class III with the intent that either the FDA would reclassify the device into a lower class or call for the premarket approval application. Due to a limited number of such applicant devices and the lack of a standard protocol, we only included applicant devices of Class I or II in our analyses. 
 \textit{Country Code} indicates the country of origin for a device manufacturer. We reduced the number of country code levels by preserving the most common ones and re-coding the others with a frequency of less than 1\% as ``Other." 
 The Center for Devices and Radiological Health (CDRH) associates each medical device with a \textit{Product Code} based upon the medical device function. For example, ultrasonic pulsed doppler devices are assigned to ``IYN," while Ultrasonic Pulsed Echo imaging devices are assigned to ``IYO." We reduced the number of product code levels for each medical specialty by preserving the most common ones and re-coding the others with a frequency of less than 5\% as ``Other Medical Specialty." 
\textit{Implantable} is a flag indicating whether a device is designed to be placed into a surgically or naturally formed cavity of the human body. 
\textit{Life Sustaining/Supporting} is a flag indicating whether a device is essential to restoring or continuing a bodily function. 

In the FDA 510(k) program, the \textit{substantial equivalence} is often evaluated based on the similarities between the predicates devices and the applicant device in terms of the composition and design (\citealt{zuckerman2014lack}). In the publicly available datasets, the information of predicate devices is not directly linked to the corresponding applicant devices. Thus, we created several variables corresponding to predicate devices, which can be classified into the following three categories. 

The first category accounts for the similarity between an applicant device and predicate devices. \textit{Number of Predicates} in this category is a count of the number of predicate devices identified for an applicant device. 
\textit{Prop. of Unmatched Medical Specialties} measures the ratio of the number of distinct medical specialties among an applicant device’s predicates that differ from the applicant device’s own specialty, divided by the total number of predicate devices. Similarly, \textit{Prop. of Unmatched Product Codes} measures the ratio of the number of distinct product codes among an applicant device's predicates that differ from the applicant device's own product code, divided by the total number of predicate devices.



The second category accounts for the age of predicate devices. \textit{Predicate Average Age} is the difference between the average year of approval of predicate devices and the application submission date of the applicant device. \textit{Predicate Median Age} is the difference between the median year of approval of predicate devices and the application submission date of the applicant device. \textit{Predicate Newest Age} indicates the difference between the year of approval of the newest predicate device and the application submission date of the applicant device. Similarly, \textit{Predicate Oldest Age} indicates the difference between the year of approval of the oldest predicate device and the application submission date of the applicant device. 

The third category accounts for the recall status of predicates. 
\textit{Number of Class 1 Recalls} is a count of the total number of Class 1 recalls among the predicate devices identified for an applicant device. \textit{Number of Class 2 Recalls} and \textit{Number of Class 3 Recalls} can be defined similarly. \textit{Variance of Recalls} is the sample variance of the number of recalls of predicate devices. For example, consider an applicant device with three predicate devices. In the first scenario, the first predicate has six recalls, while the second and third predicates have zero recalls. In the second scenario, each predicate device has two recalls. The sample variance of recalls for the first scenario is 12, while it is zero for the second scenario. 
Another interesting piece of information to consider is the timing of the recall events. For example, the importance of a recall event of a predicate device that has occurred many years prior to an applicant device's submission date may differ from a recent recall event. \textit{Weighted Recall Score} is calculated as the weighted number of recalls across all predicate devices associated with a given submission, divided by the total number of recalls for the corresponding predicate devices.
We define a time window of ten years such that a recall event is negligible if it has occurred at or more than ten years prior to the applicant's device submission date. For the other recall events, we assign weights based on the time difference. Thus, a recall event that has occurred at the submission date receives a weight of one, and a recall event that has occurred ten years before the submission date is assigned a weight of zero. For recall events that fall within this ten-year window, weights are calculated proportionally based on their distance from the submission date (e.g., a recall event that happened 4 years ago would be assigned a weight of 0.6).

Overall, we constructed 17 predictors for each applicant device, including 11 continuous measures, 3 binary indicator flags (Device Class, Implantable, and Life‑Sustaining/Supporting), and 3 multi‑level categorical variables (Medical Specialty, Product Code, and Country Code). During data preprocessing, we applied one-hot encoding to the multi-level categorical variables, resulting in 20 unique categories for Medical Specialty, 111 for Product Code, and 12 for Country Code.

\color{black}
\section{ML Models Implementation and Cross-Validation Procedures \label{Appendix-implementation}}{
In this section, we outline the key implementation details of our ML models and cross-validation (CV) strategy. All models were implemented in Python using open-source packages.

\subsection*{Data Splitting and Preprocessing}
We used a stratified random split to divide the full dataset into 70\% training and 30\% test data, ensuring the recall rate is preserved across splits. This was implemented using the \texttt{train\_test\_split()} function from the scikit-learn library.

On the training set, stratified 10-fold cross-validation was used to tune hyperparameters, implemented via the \texttt{StratifiedKFold} function also from the scikit-learn library. Continuous variables were standardized based on training data statistics.

\subsection*{ML Models}
All models were evaluated using mean AUC over 10-fold CV, repeated 10 times with shuffled folds.

\begin{itemize}
  \item \textbf{Logistic Regression (Lasso and Ridge):} Implemented using \texttt{sklearn.linear\_model.LogisticRegression} with \texttt{liblinear} solver. Penalty type (\texttt{l1} for Lasso, \texttt{l2} for Ridge) and regularization parameter \texttt{C} were tuned.

  \item \textbf{Decision Tree Classifier:} Implemented using \texttt{sklearn.tree.DecisionTreeClassifier}. Hyperparameters \texttt{min\_samples\_split} and \texttt{max\_depth} were tuned.

  \item \textbf{Random Forest Classifier:} Implemented using \texttt{sklearn.ensemble.RandomForestClassifier}. Hyperparameters \texttt{n\_estimators} and \texttt{max\_depth} were tuned. For improved performance and to enable parallelism, we set \texttt{max\_features = sqrt} and \texttt{n\_jobs = -1}.

  \item \textbf{Gradient Boosting Classifier:} Implemented using \texttt{sklearn.ensemble.GradientBoosting\discretionary{}{}{}Classifier}. Hyperparameters \texttt{n\_estimators} and \texttt{learning\_rate} were tuned. We fixed \texttt{max\_depth = 4} and \texttt{subsample = 0.8}.

  \item \textbf{Cox Proportional Hazards Model (Lasso and Ridge):} Implemented using \texttt{sklearn\_survival.CoxnetSurvivalAnalysis}. Regularization parameter \texttt{alpha} was tuned. For Lasso, \texttt{l1\_ratio = 1.0} and for Ridge, \texttt{l1\_ratio = 1e-5}.
\end{itemize}
}

\section{Robustness Checks for Censoring Effect \label{Appendix-Censoring}}{
Recall that in our ML models, the primary outcome is a binary recall event indicating whether an applicant device had at least one recall between its clearance date and the end of our observation window. We include all devices cleared from 2008 to 2020 and we track their recall events until the end of 2021. This could create a censoring effect for the most recent applicant devices. That is, the devices approved at dates closer to the end of the observation period may have a lower chance of experiencing a recall than those approved at the beginning of the observation period. In this section, we employ two robustness checks to assess the impact of censoring.

In our first method, we apply survival analysis techniques that are commonly used in the literature in the presence of censored outcomes. The Cox proportional hazards model is a widely used survival analysis technique known for its semi-parametric nature and interpretability. It models the hazard rate for an event as a linear combination of covariates, allowing the coefficients to be interpreted in terms of hazard ratios. However, standard Cox models can struggle with correlated feature spaces, which leads to numerical instability when inverting covariance matrices. To address this and avoid overfitting, we employ regularization techniques by incorporating Lasso and Ridge penalties into the Cox model.

We train two regularized Cox models with Ridge and Lasso penalties using scikit-survival, which is a Python module for survival analysis built on top of the scikit-learn library. Penalty parameters are optimized through a 10-fold cross-validation procedure based on the area under the curve (AUC) metric. Table~\ref{tableEC-CoxAUC} reports the predictive performance of the regularized Cox models on unseen 510(k) applicant devices. The cross-validated AUC values for the two Cox models range from 0.74 to 0.77. Among them, the Cox model with Ridge penalty demonstrates a higher AUC with lower variance. However, it still underperforms compared to gradient boosting, which remains our selected model for recall risk prediction. 


\begin{table}[b]
\centering \small
\caption{Performance comparison of survival models}
\label{tableEC-CoxAUC}
\begin{tabular}{@{}l c c@{}} 
\toprule
\textbf{Model} & \textbf{CV-AUC (95\% CI)} & \textbf{OOS-AUC (95\% CI)} \\
\midrule
Cox with Ridge penalty & 0.77 (0.73, 0.81) & 0.76 (0.75, 0.77) \\
Cox with Lasso penalty & 0.74 (0.70, 0.78) & 0.74 (0.72, 0.75) \\

\bottomrule
\end{tabular}
\end{table}

Our second method involves an incremental analysis to systematically assess the impact of censoring. In this approach, we construct train and test datasets where devices have sufficient follow-up time to capture potential recall events. 
This setup enables us to analyze how varying levels of censoring in data affect model performance. Figure~\ref{figEC-RecallTime} shows the distribution of recall time for all devices in our dataset that have been recalled. The bars show the absolute percentage of recalls after years from submission, and the red curve shows the cumulative recall percentage up to different years after submission. The figure shows that a substantial portion of recalls occurs relatively early, where 71.9\% of recalls happen up to 4 years after submissions.

In our incremental analysis, we consider follow-up thresholds of at least 9, 8, 7, 6, 5, and 4 years, allowing us to systematically assess how varying levels of censoring in data impact model performance. For example, a threshold of 9 years in this analysis means that we only include devices that have had at least 9 years of follow-up in our data. According to Figure~\ref{figEC-RecallTime}, 95.6\% of these devices should have already experienced a recall event (if applicable), which ensures minimal censoring in the dataset. 
By gradually reducing the follow-up threshold, we assess whether training with progressively censored data impacts model predictions. Table~\ref{tableEC-IncrementalAUC} reports the CV-AUC of our selected gradient boosting model across different follow-up thresholds. \textcolor{black}{The CV-AUC values range from 0.75 to 0.76 and the OOS-AUC values range from 0.74 to 0.76, indicating that the model maintains stable performance across varying levels of data censoring.} Comparing these results to the model's AUC using all data, we find that our ML model is robust to potential censoring effects introduced by more recent device approvals. Our additional analysis further confirms that the main findings regarding important variables are also robust to censoring (see Appendix \ref{Appendix-VarSig}).

\begin{figure}[tb]
\centerline{\includegraphics[scale=.45]{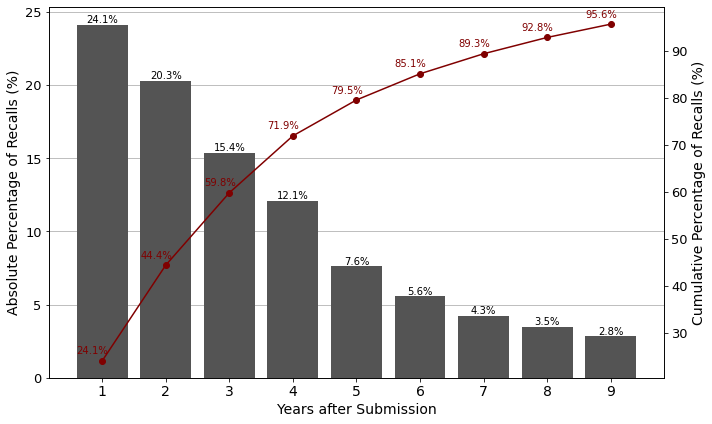}}
\caption{Distribution of recall time}
\label{figEC-RecallTime}
\end{figure}


\begin{table}[tb]
\centering \small
\caption{Area under the curve across follow-up times}
\label{tableEC-IncrementalAUC}
\begin{tabular}{@{}l c c@{}}
\toprule
\textbf{Years After Submission} & \textbf{CV-AUC (95\% CI)} & \textbf{OOS-AUC (95\% CI)} \\
\midrule
$\; \ge \;$9 & 0.75 (0.70, 0.80) & 0.74 (0.72, 0.76) \\
$\; \ge \;$8 & 0.75 (0.71, 0.80) & 0.74 (0.73, 0.76) \\
$\; \ge \;$7 & 0.75 (0.71, 0.80) & 0.76 (0.75, 0.78) \\
$\; \ge \;$6 & 0.75 (0.71, 0.80) & 0.76 (0.75, 0.78) \\
$\; \ge \;$5 & 0.75 (0.71, 0.79) & 0.76 (0.75, 0.78) \\
$\; \ge \;$4 & 0.76 (0.73, 0.80) & 0.75 (0.74, 0.77) \\
\bottomrule
\end{tabular}
\end{table}

\textcolor{black}{Based on our robustness checks, we find that the predictive performance of our selected model as well as the identification of key variables remain stable across different levels of data censoring. Thus, it provides a reliable approach for recall risk prediction. }
}

\section{Robustness Check for Variable Significance} \label{Appendix-VarSig}
In this section, we aim to validate our insights on the important variables correlated with recall risk by examining the important variables in a logistic regression with Lasso regularization (Log Reg with Lasso). Our model comparison results (see Table~\ref{table-AUC}) indicated that Log Reg with Lasso performs comparably to our primary model (gradient boosting), making it a useful alternative for validating our key findings. Additionally, we use Cox with Lasso (see Table~\ref{tableEC-CoxAUC}) as another alternative to check whether our key findings are robust to potential censoring effects.

We directly assess the significance of the predictors retained in both models. The standard inference tools (e.g., $p$‐values) are not directly applicable to regularized models. To address this, we adopt a post‐Lasso (refit) approach with sample‐splitting (\citealt{wasserman2009high}). Specifically, we fit each model on the training set to identify variables with non-zero coefficients, and then fit a logistic/Cox model with no regularization on the testing set using only these selected variables.
Tables~\ref{tableEC2} and \ref{tableEC3} report the coefficients, standard errors, and p‐values from the refitted logistic regression and Cox proportional hazard models. According to Table~\ref{tableEC2},  we find that all variables identified as important in the SHAP analysis of the selected model (gradient boosting) remain statistically significant, except for Predicate Median Age, Oldest Age, and Variance of Recalls. Although the p‐value for Predicate Median Age is not statistically significant at the 0.05 level (p = 0.066), the variable still appears to be an important predictor  and exhibits a negative association with the log-odds of recall. While Predicate Oldest Age is among the selected variables in Log Reg with Lasso, its p‐value is not statistically significant. We also note that the Variance of Recalls was not retained by the Log Reg with Lasso model. Similar results are observed in Table~\ref{tableEC3} corresponding to the Cox with Lasso model. 

\begin{table}[tb]
\centering \small
\caption{Output summary of the logistic regression model with selected variables by Log Reg with Lasso}
\label{tableEC2}
\begin{tblr}{
  hline{1-2,13,15} = {-}{},
  cell{15}{1} = {c=4}{},
  cell{16}{1} = {c=4}{}
}
\textbf{Independent Variable}        & \textbf{Coefficient} & \textbf{Standard Error} & \textbf{$p$-value} \\
Weighted Recall Score               & 0.892       & 0.112          & $0.000^{*}$ \\
Num. of Class 1 Recalls             & -0.153      & 0.156          & 0.326         \\
Num. of Class 2 Recalls             & 0.060       & 0.020          & $0.002^{*}$ \\
Predicate Median Age                & -0.022      & 0.012          & $0.066$   \\
Predicate Newest Age                & -0.030      & 0.010          & $0.003^{*}$ \\
Predicate Oldest Age                & -0.001      & 0.007          & 0.910         \\
Life Sustaining/Supporting          & 0.906       & 0.203          & $0.000^{*}$ \\
Device Class                        & -0.880      & 0.217          & $0.000^{*}$ \\
Implantable                         & -0.054      & 0.174          & 0.758         \\
Prop. of Unmatched Specialties      & -0.369      & 0.205          & 0.072   \\
Prop. of Unmatched Product Codes    & -0.056      & 0.124          & 0.653         \\
Country Codes                       & Yes         &                &               \\
Product Codes                       & Yes         &                &               \\
{\footnotesize  
Note: Significance code $^{*}$ indicates $p\,<\,$0.05}.    &   &     
\end{tblr}
\end{table}

\begin{table}[tb]
\centering \small
\caption{Output summary of the Cox proportional hazard model with variables selected by Cox with Lasso}
\label{tableEC3}
\begin{tblr}{
  hline{1-2,10,12} = {-}{},
  cell{15}{1} = {c=4}{},
  cell{16}{1} = {c=4}{}
}
\textbf{Independent Variable}        & \textbf{Coefficient} & \textbf{Standard Error} & \textbf{$p$-value} \\
Weighted Recall Score               & 0.893 & 0.069 & $0.000^{*}$ \\
Num. of Class 1 Recalls             & 0.182 & 0.058 & $0.002^{*}$ \\
Num. of Class 2 Recalls             & 0.067 & 0.009 & $0.000^{*}$ \\
Predicate Median Age                & $-0.004$ & 0.009 & 0.626 \\
Predicate Newest Age                & $-0.033$ & 0.008 & $0.000^{*}$ \\
Predicate Oldest Age                & 0.008 & 0.005 & 0.109 \\
Life-Sustaining/Supporting          & 0.902 & 0.098 & $0.000^{*}$ \\
Device Class                        & $-0.520$ & 0.215 & $0.016^{*}$ \\
Country Codes                       & Yes   &       &       \\
Product Codes                       & Yes   &       &       \\
{\footnotesize Note: Significance code $^{*}$ indicates $p\,<\,0.05$}. & & & \\
\end{tblr}
\end{table}

Overall, the consistent direction and significance of major predictors (Weighted Recall Score, Number of Class 2 Recalls, Predicate Newest Age, Device Class, Life‐Sustaining/Supporting) across both models strengthen confidence in our main findings on the significant predictors of recall risk. Regarding the Predicate Median Age, Predicate Oldest Age, and Variance of Recalls, although they help the selected model to have a better prediction of recall risk, the results suggest that one needs to be careful about their association with recall risk in general.

\section{Sensitivity Analysis of the Key Input Parameters of the Optimization Model}
\label{ImpactofInput}

We investigate the impact of input parameters for our optimization model on various metrics, including the acceptance and rejection rates of both safe and unsafe devices, as well as the FDA's workload (currently stands at 100\% as all devices are evaluated by the FDA's committees). We also measure the percentage point (pp) improvement over the FDA’s current practice, which is defined as the difference between the FDA’s current recall rate and the recall rate under our policy. This information can assist decision-makers in selecting the right input parameters that align with their criteria. The results for the training and test sets are summarized in Tables~\ref{tableEC-SensitivityTrain} and \ref{tableEC-SensitivityTest}. In these tables, each input parameter is considered at three levels: low (L), medium (M), and high (H). For the parameters $\xi^{ru}$ and $\xi^{as}$, we have $L = 0.3$, $M = 0.5$, and $H = 0.7$. For the parameter $p$, the values are $L = 0.4$, $M = 0.6$, and $H = 0.8$. We note that the ``N/A" values in the last three rows indicate that the corresponding input parameters render the problem infeasible. The infeasibility of these cases stems from the fulfillment of condition (c) in Theorem \ref{thm2}. 

As an example, consider the case where $\xi^{ru} = M$, $\xi^{as} = L$, and $p = M$ (i.e., M-L-M combination) in Table~\ref{tableEC-SensitivityTrain}, which presents the metric values for the training set. We observe that using our policy results in rejecting 50.2\% of unsafe devices and accepting 31.0\% of safe devices. Among the accepted devices, 5.0\% would experience a future recall, while 14.5\% of the rejected devices would face no future recall. 
Accordingly, this policy leads to a 5.2 percentage point improvement (i.e., 50.2\% improvement) in the recall rate compared to the FDA's current practice, which is 10.3\%. Additionally, it results in a 46.6\% reduction in the workload of the FDA's committees, as only 53.4\% of the devices will be forwarded to the FDA’s committees for in-depth evaluation, while the remaining 46.6\% will be automatically accepted or rejected.

We observe substantial gains in certain scenarios, such as the L-L-L combination in Table~\ref{tableEC-SensitivityTest}, which results in a remarkable 60.1\% workload reduction and a significant 6 percentage point improvement (i.e., 58.6\% improvement). However, it is essential to note a vital caveat. Some combinations that offer significant improvements in workload reduction and recall rate also lead to an unreasonably high rate of rejection of safe devices. This phenomenon arises in the specific setting of low and high thresholds that, in certain cases, reject a substantial proportion of safe devices. Therefore, the selection of input parameters must strike a delicate balance between reducing workload and maintaining an acceptable rate of rejection for safe devices. 

\begin{table} [tbh]
\centering \footnotesize
\caption{Comparative analysis of key metrics (percentages) across combinations of input parameters in the training set}
\label{tableEC-SensitivityTrain}
\begin{tblr}{
hline{1-2,29} = {-}{},
cell{29}{1} = {c=10}{}
}
$\xi^{ru}$ & $\xi^{as}$ & $\rho$& {Reject \\Unsafe} & {Accept \\Safe} & {Accept \\Unsafe} & {Reject\\Safe} & {Reject \\Rate} & Workload & {Absolute \\Imp. (pp)} \\
L & L & L & 66.2 & 31.0 & 5.0  & 28.1 & 32.1 & 39.6 & 6.8 \\
L & L & M & 41.3 & 31.0 & 5.0  & 8.3  & 11.7 & 60.0 & 4.2 \\
L & L & H & 30.1 & 31.0 & 5.0  & 3.5  & 6.2 & 65.4 & 3.1 \\ 
L & M & L & 42.2 & 51.2 & 15.6 & 8.9  & 12.3 & 40.1 & 4.4 \\
L & M & M & 30.1 & 51.2 & 15.6 & 3.5  & 6.2 & 46.2 & 3.1 \\
L & M & H & 30.1 & 51.2 & 15.6 & 3.5  & 6.2 & 46.2 & 3.1 \\
L & H & L & 30.1 & 70.7 & 32.2 & 3.5  & 6.2 & 27.1 & 3.1 \\
L & H & M & 30.1 & 70.7 & 32.2 & 3.5  & 6.2 & 27.1 & 3.1 \\
L & H & H & 30.1 & 70.7 & 32.2 & 3.5  & 6.2 & 27.1 & 3.1 \\
M & L & L & 66.2 & 31.0 & 5.0  & 28.1 & 32.1 & 39.6 & 6.8 \\
M & L & M & 50.2 & 31.0 & 5.0  & 14.5 & 18.2 & 53.4 & 5.2 \\
M & L & H & 50.2 & 31.0 & 5.0  & 14.5 & 18.2 & 53.4 & 5.2 \\
M & M & L & 50.2 & 51.2 & 15.6 & 14.5 & 18.2 & 34.2 & 5.2 \\
M & M & M & 50.2 & 51.2 & 15.6 & 14.5 & 18.2 & 34.2 & 5.2 \\
M & M & H & 50.2 & 51.2 & 15.6 & 14.5 & 18.2 & 34.2 & 5.2 \\
M & H & L & 50.2 & 70.7 & 32.2 & 14.5 & 18.2 & 15.1 & 5.2 \\
M & H & M & 50.2 & 70.7 & 32.2 & 14.5 & 18.2 & 15.1 & 5.2 \\
M & H & H & 50.2 & 70.7 & 32.2 & 14.5 & 18.2 & 15.1 & 5.2 \\
H & L & L & 71.1 & 31.0 & 5.0  & 32.8 & 36.8 & 34.9 & 7.3 \\
H & L & M & 71.1 & 31.0 & 5.0  & 32.8 & 36.8 & 34.9 & 7.3 \\
H & L & H & 71.1 & 31.0 & 5.0  & 32.8 & 36.8 & 34.9 & 7.3 \\
H & M & L & 71.1 & 51.2 & 15.6 & 32.8 & 36.8 & 15.7 & 7.3 \\
H & M & M & 71.1 & 51.2 & 15.6 & 32.8 & 36.8 & 15.7 & 7.3 \\
H & M & H & 71.1 & 51.2 & 15.6 & 32.8 & 36.8 & 15.7 & 7.3 \\
H & H & L & ``N/A" & ``N/A" & ``N/A" & ``N/A" & ``N/A" & ``N/A" & ``N/A" \\
H & H & M & ``N/A" & ``N/A" & ``N/A" & ``N/A" & ``N/A" & ``N/A" & ``N/A" \\
H & H & H & ``N/A" & ``N/A" & ``N/A" & ``N/A" & ``N/A" & ``N/A" & ``N/A" \\
{\footnotesize Note: The recall rate of the FDA's current practice in the training set is 10.3\%.} \\
\end{tblr}
\end{table}

\begin{table} [tb]
\centering \footnotesize
\caption{Comparative analysis of key metrics (percentages) across combinations of input parameters in the testing set}
\label{tableEC-SensitivityTest}
\begin{tblr}{
hline{1-2,29} = {-}{},
cell{29}{1} = {c=10}{}
}
$\xi^{ru}$ & $\xi^{as}$ & $\rho$& {Reject \\Unsafe} & {Accept \\Safe} & {Accept \\Unsafe} & {Reject\\Safe} & {Reject \\Rate} & Workload & {Absolute \\Imp. (pp)} \\
L & L & L & 58.6 & 30.7 & 8.9  & 28.6 & 31.7 & 39.9 & 6.0 \\
L & L & M & 32.9 & 30.7 & 8.9  & 9.7 & 12.1 & 59.5 & 3.4 \\
L & L & H & 19.9 & 30.7 & 8.9  & 4.7 & 6.3 & 65.3 & 2.0 \\
L & M & L & 33.7 & 50.0 & 22.8 & 10.1 & 12.6 & 40.3 & 3.5 \\
L & M & M & 19.9 & 50.0 & 22.8 & 4.7 & 6.3 & 46.6 & 2.0 \\
L & M & H & 19.9 & 50.0 & 22.8 & 4.7 & 6.3 & 46.6 & 2.0 \\
L & H & L & 19.9 & 70.1 & 39.5 & 4.7 & 6.3 & 26.8 & 2.0 \\
L & H & M & 19.9 & 70.1 & 39.5 & 4.7 & 6.3 & 26.8 & 2.0 \\
L & H & H & 19.9 & 70.1 & 39.5 & 4.7 & 6.3 & 26.8 & 2.0 \\
M & L & L & 58.6 & 30.7 & 8.9  & 28.6 & 31.7 & 39.9 & 6.0 \\
M & L & M & 42.8 & 30.7 & 8.9  & 15.4 & 18.2 & 53.3 & 4.4 \\
M & L & H & 42.8 & 30.7 & 8.9  & 15.4 & 18.2 & 53.3 & 4.4 \\
M & M & L & 42.8 & 50.0 & 22.8 & 15.4 & 18.2 & 34.6 & 4.4 \\
M & M & M & 42.8 & 50.0 & 22.8 & 15.4 & 18.2 & 34.6 & 4.4 \\
M & M & H & 42.8 & 50.0 & 22.8 & 15.4 & 18.2 & 34.6 & 4.4 \\
M & H & L & 42.8 & 70.1 & 39.5 & 15.4 & 18.2 & 14.8 & 4.4 \\
M & H & M & 42.8 & 70.1 & 39.5 & 15.4 & 18.2 & 14.8 & 4.4 \\
M & H & H & 42.8 & 70.1 & 39.5 & 15.4 & 18.2 & 14.8 & 4.4 \\
H & L & L & 63.8 & 30.7 & 8.9  & 33.4 & 36.6 & 35.0 & 6.5 \\
H & L & M & 63.8 & 30.7 & 8.9  & 33.4 & 36.6 & 35.0 & 6.5 \\
H & L & H & 63.8 & 30.7 & 8.9  & 33.4 & 36.6 & 35.0 & 6.5 \\
H & M & L & 63.8 & 50.0 & 22.8 & 33.4 & 36.6 & 16.3 & 6.5 \\
H & M & M & 63.8 & 50.0 & 22.8 & 33.4 & 36.6 & 16.3 & 6.5 \\
H & M & H & 63.8 & 50.0 & 22.8 & 33.4 & 36.6 & 16.3 & 6.5 \\
H & H & L & ``N/A" & ``N/A" & ``N/A" & ``N/A" & ``N/A" & ``N/A" & ``N/A" \\
H & H & M & ``N/A" & ``N/A" & ``N/A" & ``N/A" & ``N/A" & ``N/A" & ``N/A" \\
H & H & H & ``N/A" & ``N/A" & ``N/A" & ``N/A" & ``N/A" & ``N/A" & ``N/A" \\
{\footnotesize Note: The recall rate of the FDA's current practice in the testing set is 10.3\%.}
\end{tblr}
\end{table}

\section{Robustness Check for Selection Bias} \label{EC:SelectionBias}

As noted in Section~\ref{DataCollection}, our dataset, constructed from publicly available 510(k) records, includes only those devices that were granted market authorization by the FDA. Generally, there are two reasons a 510(k) submission does not proceed to clearance. The first is a Refusal to Accept (RTA), an administrative rejection based on application completeness. Because applicants may resubmit corrected applications, these devices typically appear in our dataset once clearance is eventually granted. The second is a Not Substantially Equivalent (NSE) determination, which may result either from predicate ineligibility (e.g., differences in intended use or technological characteristics) or from failure during substantive review (i.e., the device passed the eligibility check but failed performance testing).

The absence of NSE data does not affect model training, as recalls are post-market events and rejected devices therefore cannot be used for training our ML model. However, access to NSE determinations arising from failure during substantive review (as opposed to predicate ineligibility) could be informative for model evaluation. Internal analyses by the CDRH indicate that roughly 5\% of all 510(k) submissions are ultimately classified as NSE (\citealt{CDRH}). While this small fraction of rejections mitigates concerns about selection bias, it does not eliminate them. To address this limitation, we conduct a robustness check by augmenting our dataset with synthetic devices that simulate devices rejected by the FDA, and then re-evaluating our policy performance in terms of recall rate percentage improvement and rejection rate of safe devices.

In particular, we perform a sensitivity analysis by generating synthetic applicant devices rejected during the substantive review. To do this, we introduce a parameter $\nu$ that represents a risk quantile, which allows us to control the risk profile of synthetic devices. Given $\nu$ and the estimated recall risk from our chosen model, we first sample 5\% of the original dataset from devices whose estimated risk exceeds $\nu \in \{0.60, 0.75, 0.95 \}$. We then label these sampled devices as synthetic FDA‐rejected cases. 
To reflect the assumption that FDA rejections for performance deficiencies correspond to actual safety risks, we assign a recall outcome to all synthetic devices. Consistent with our methodology, we do not use these synthetic devices to retrain our selected ML model, as real-world rejected devices lack the market outcomes required for training. Instead, we add these synthetic devices to the testing set and apply our existing policy, derived solely from cleared devices, to evaluate its performance. Consequently, FDA-rejected devices that are deferred by our policy for further evaluation by an FDA committee are assumed to end up receiving a rejection decision in our analysis. 

Figure~\ref{figEC4} depicts two key performance metrics in our representative setting (using the L-L-M combination), including the rejection rate of safe devices and the recall rate percentage improvement. As expected by the design of our analysis, we observe that the rejection rate of safe devices remains constant at the baseline level of 9.7\%. 
Concurrently, the recall rate percentage improvement remains substantial, ranging from 30.8\% to 23.9\% across the different risk profiles. The improvement is highest (30.8\%) when we simulate a targeted rejection of only the highest-risk devices ($\nu=0.90$), which are easier for our policy to identify. As we broaden the scope to include lower-risk rejected devices ($\nu=0.60$), the improvement decreases slightly to 23.9\% but remains significant compared to the baseline FDA process. This robustness suggests that relying on cleared devices for model training, a structural necessity given the lack of market outcomes for rejected applications, does not undermine the general effectiveness of our proposed policy. 
\begin{figure}[tb]
\centerline{\includegraphics[scale=.5]{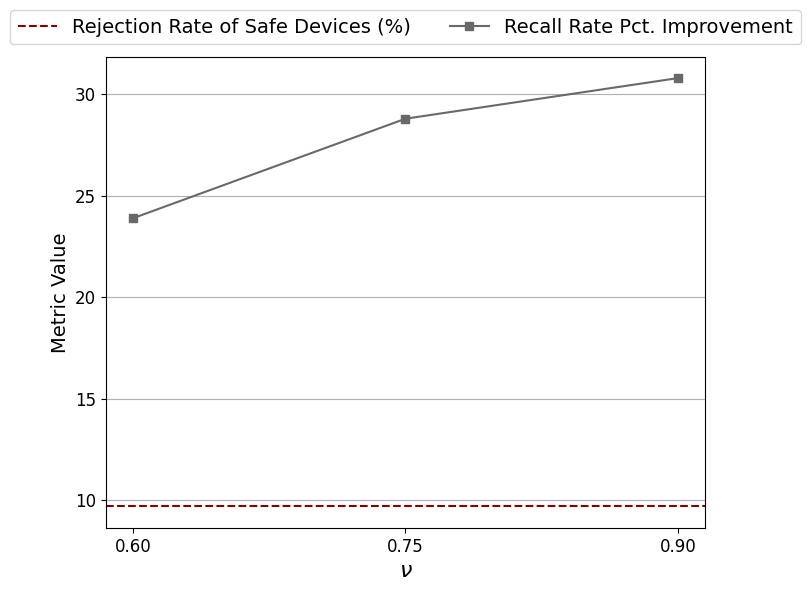}}
\caption{Impact of unobserved (FDA-rejected) devices on key metrics}
\label{figEC4}
\end{figure}

\section{Post-Hoc Analysis} \label{Post-Hoc Analyses} 
In this section, we delve deeper into assessing the performance of our proposed policy via post-hoc analyses. We start by examining the characteristics of the deferred medical devices based on some of the top predictors of recall risk. The results are summarized in Table~\ref{table6}. In this table, we also compare devices that are deferred by our proposed policy for more in-depth evaluation to those that have been correctly accepted or rejected. It is evident that the risk estimates are higher for unrecalled devices deferred by our policy compared to those that have been correctly accepted. Similarly, the risk estimates are lower for recalled devices deferred by the policy compared to those that have been correctly rejected. For unrecalled devices, deferred cases have modestly elevated recall histories and slightly newer predicates compared to accepted ones. For recalled devices, deferred cases exhibit lower recall histories and slightly older predicates than those that were confidently rejected.  These patterns suggest that deferred devices fall into a borderline zone where risk signals and related features raise some concern, but not strongly enough to support a confident accept or reject decision. This supports the need for additional human review and the design of our policy to defer hard cases for further evaluation.

\begin{table}
\centering \small
\caption{Comparison of devices correctly evaluated and deferred by our policy in the testing set}
\label{table6}
\begin{tblr}{
  cell{1}{2} = {c=3}{},
  cell{1}{6} = {c=3}{},
  hline{1,4,11} = {-}{},
  hline{2} = {2-8}{},
  hline{3} = {2,4,6,8}{},
}
                              & Unrecalled Devices &  &             &  & Recalled Devices &  &             \\
                              & Accepted           &  & Deferred    &  & Rejected         &  & Deferred    \\
                              & Mean (SD)          &  & Mean (SD)   &  & Mean (SD)        &  & Mean (SD)   \\
Risk Estimate                 & 0.04 (0.01)        &  & 0.10 (0.03) &  & 0.32 (0.14)      &  & 0.10 (0.03) \\
Predicate Median Age          & 4.74 (4.20)        &  & 3.81 (3.33) &  & 3.49 (2.84)      &  & 3.86 (3.44) \\
Predicate Newest Age          & 4.57 (5.26)        &  & 3.63 (4.50) &  & 2.84 (3.64)      &  & 3.46 (4.31) \\
Predicate Oldest Age          & 7.86 (6.70)        &  & 7.08 (6.32) &  & 7.23 (6.55)      &  & 7.41 (6.86) \\
Num. of Class 2 Recalls       & 0.07 (0.56)        &  & 0.26 (1.00) &  & 2.84 (5.04)      &  & 0.44 (2.99) \\
Weighted Num. of Recalls      & 0.02 (0.10)        &  & 0.09 (0.23) &  & 0.57 (0.38)      &  & 0.11 (0.26) \\
Variance of Recalls           & 0.00 (0.05)        &  & 0.01 (0.10) &  & 0.39 (2.87)      &  & 0.02 (0.22) 
\end{tblr}
\end{table}

Table~\ref{table7} presents a comparison of devices accepted by our proposed policy and those accepted based on the FDA's current practice in the testing set. As our policy does not directly diagnose the deferred devices, this analysis is conducted under the assumption that the deferred devices are evaluated following the FDA's current practice. We observe that the risk estimate for devices accepted by our proposed policy is relatively similar to that of the FDA’s accepted devices (0.08 vs. 0.10). This indicates that both approaches accept devices with comparably low estimated risk; however, the predicate characteristics differ. The age of the latest-approved predicate for devices accepted under our policy is slightly higher compared to devices accepted under the FDA's current practice, while the age of the earliest-approved predicate is relatively the same. This shows a balance between leveraging proven safety records and embracing new technology. 
Furthermore, the number of class 2 recalls, weighted number of recalls, and variance of predicates' recalls for devices accepted by our policy is significantly lower than those accepted by FDA's current practice. This is a substantial difference and indicates that our policy has the tendency to accept devices with predicates with fewer historical recalls, which aligns with the existing literature suggesting the correlation between the chance of recall and the number of recalls for predicates. 

\begin{table}
\centering \small
\caption{Comparison of accepted devices by our policy and current practice in the testing set}
\label{table7}
\begin{tblr}{
  hline{1,3,10} = {-}{},
  hline{2} = {2,4}{},
}
                              & Our Policy       &  & Current Practice     \\
                              & Mean (SD)        &  & Mean (SD)            \\
Risk Estimate                 & 0.08 (0.04)      &  & 0.10 (0.09)          \\
Predicate Median Age          & 4.11 (3.66)      &  & 4.04 (3.58)          \\
Predicate Newest Age          & 3.92 (4.76)      &  & 3.78 (4.65)          \\
Predicate Oldest Age          & 7.35 (6.48)      &  & 7.33 (6.48)          \\
Num. of Class 2 Recalls       & 0.21 (1.16)      &  & 0.49 (2.57)          \\
Weighted Num. of Recalls      & 0.07 (0.20)      &  & 0.12 (0.28)          \\
Variance of Recalls           & 0.01 (0.10)      &  & 0.06 (1.42)          
\end{tblr}
\end{table}

\section{A Comparison of Human-Algorithm and ML-Only Policies} \label{EC:Comparison}

In this section, we compare the performance of our policy, which is a combined human-algorithm approach, with that of an ML-only policy that is solely based on non-human decisions. We conduct this comparison in two distinct parts: first, an analysis of operational trade-offs under a fixed set of preferences, and second, a more robust comparison of the policies' entire performance frontiers.

The recall rate percentage improvement relative to the FDA's current practice is an important metric. However, the overall rejection rate is also a critical operational factor for the FDA, as adopting a policy that results in a significant number of submission rejections may not be practical. Accordingly, we first compare the two policies based on these two metrics under a fixed set of preferences and a conservative evaluation. We solve our \textcolor{black}{Primary Problem} using the same set of parameters ($\lambda = 0.8$, $\xi^{ru} = 0.3$, and $\xi^{as} = 0.3$) while varying the threshold on the FDA's workload $\rho$ from zero to one. The solution at $\rho=0$ represents the optimal ML-only policy under these exact constraints, as a zero workload requirement forces the optimizer to find the best single-threshold solution. The solutions for $\rho > 0$ show the marginal benefit of the deferral option in our human-algorithm policy. This approach ensures both policies satisfy the same minimum performance constraints and use the same priority-weighting ($\lambda$).

Figure~\ref{figEC-Comp1} shows the recall rate percentage improvement versus the overall rejection rate, with the FDA workload annotated by text on the curve. Under this fixed set of preferences, the ML-only policy ($\rho=0$) yields a 90.5\% improvement in recall rate, but does so by rejecting an impractical 71.6\% of all submissions. Our policy navigates this trade-off more effectively. The red curve shows that our policy provides an operational lever, allowing a decision-maker to trade this high recall rate for a more balanced policy.
For example, our policy can reduce the rejection rate to 12.1\% and reduce the FDA's workload by 40.5\%, while still achieving a 32.9\% improvement in recall rate. This demonstrates that even in a conservative setting, our policy provides critical flexibility.

\begin{figure} [tb] 
    \centering \footnotesize
    \includegraphics[width=0.55\linewidth]{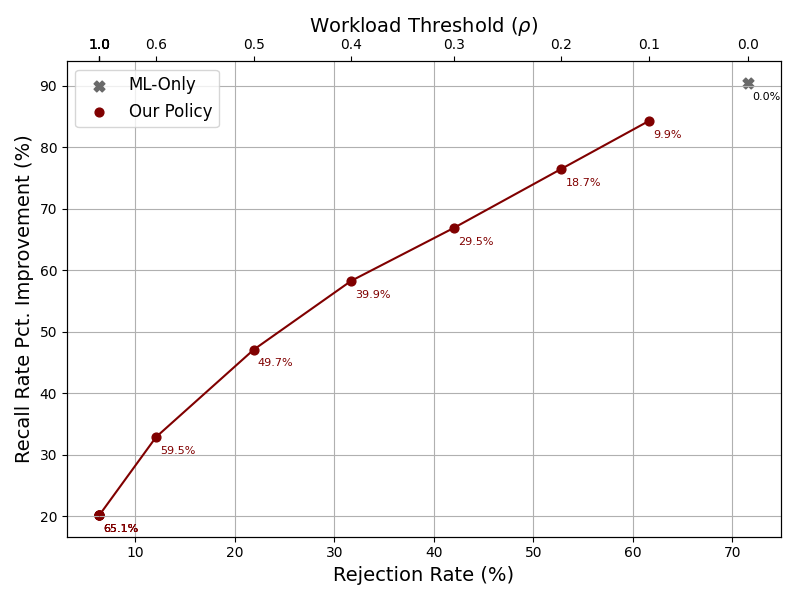}
    \caption{Comparison of the performance of our policy vs the ML-only policy in the testing set under  ($\lambda = 0.8$, $\xi^{ru} = 0.3$, and $\xi^{as} = 0.3$). The annotated text on the curve indicates the corresponding FDA's workload.} 
    \label{figEC-Comp1}
\end{figure}


While the first analysis shows that our policy provides a valuable operational trade-off, the other question is whether our policy is fundamentally superior compared to an ML-only policy. To answer this, we can compare the entire performance frontier of both policies by varying the priority-weighting parameter ($\lambda$). However, conducting this frontier-level comparison in the conservative setting is uninformative. Unlike the ML-only policy, which does not result in deferring any applicant devices, our policy results in deferring some proportion of applicant devices to the FDA committee for more in-depth evaluation in a setting with additional a priori information about the risk of the applicant device and lower workload. Consequently, the performance of our policy partly depends on the performance of the FDA committee to evaluate the deferred devices. In particular, for any risk thresholds $(\ell, h)$ selected by our policy, there exists a corresponding threshold for the ML-only policy by setting $t = h$. With this construction, both policies yield the same acceptance rate for unsafe devices and the same rejection rate for safe devices. Devices with estimated risk at or above $h$ are rejected under both policies, as the ML-only threshold is the same as the rejection boundary of our policy. Devices with risk below $\ell$ are accepted by both approaches since the single threshold $h$ is equal to or greater than $\ell$. For devices with risk between $\ell$ and $h$, the ML-only policy accepts the device, whereas our policy defers the decision to the FDA. However, because our dataset contains only devices that were cleared (accepted) by the FDA, all deferred devices will be accepted as well. 

Accordingly, we move to a non-conservative setting by modeling the added benefits of our policy. These benefits include providing additional information for deferred devices and reducing the FDA’s overall workload. This can potentially allow the committee to better determine the review depth and level of scrutiny required for each applicant device; for example, riskier devices may undergo more rigorous evaluations, while the reduced workload allows committees to focus their expertise on the most complex cases. We assume our policy benefits the FDA by providing additional information and reducing workload, allowing the committee to better evaluate deferred cases. We model the probability that the FDA's committees will fail to detect an unsafe deferred device as follows:
$$  \mathcal{L}(f(X), k) = 2 \left( 1  - \frac{1}{1+\exp \left( -k (f(X)- \hat{\ell}) \right)} \right),$$
where $f(X)$ is the estimated risk, $\hat{\ell}$ is the low threshold obtained by our policy, and parameter $k \ge 0$ is a scalar where higher values of $k$ correspond to improved performance of the FDA's committees in detecting unsafe devices. When $k=0$, this probability is equal to $1$ by our design for any unsafe device that has been deferred. This implies that $k=0$ is the baseline that matches the FDA's current practice in evaluating deferred devices without supplementary information.

Figure~\ref{fig-ProbFail} illustrates the relationship between the predicted risk and the probability of failing to reject an unsafe device for 100 random samples, ordered by their predicted risk. As can be seen, the probability of failure decreases as the predicted risk of the deferred device increases. When $k = 0$, the FDA's committees will fail to reject an unsafe device with a probability of 1. As $k$ increases, this probability decreases. This observation aligns with our intuition that unsafe devices appearing less risky are more challenging for the FDA's committees to detect. We acknowledge that this non-conservative analysis is stylized and relies on the specific parametric form of $\mathcal{L}(f(X), k)$ to model the potential benefits of deferral. Its purpose is to illustrate the potential for our policy to create a dominant frontier under improved expert performance, rather than to definitively model the complex operational trade-offs of time or resources.

\begin{figure} [tbh]
    \centering
    \includegraphics[width=0.5\linewidth]{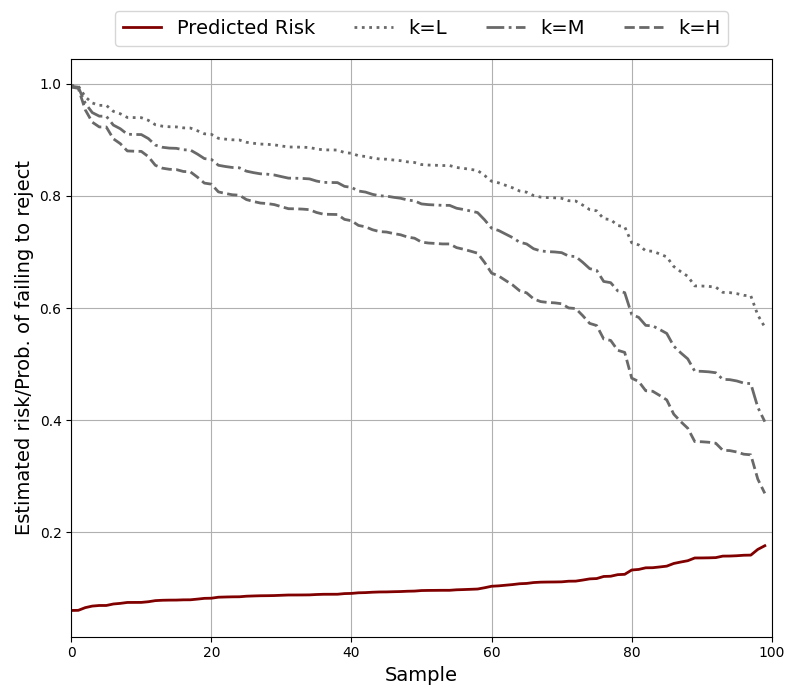}
    \caption{Predicted risk and probability of failing to reject unsafe deferred devices}
    \label{fig-ProbFail}
\end{figure}

Using this non-conservative framework, we trace the entire performance frontier for both policies by varying $\lambda \in [0.1,0.9]$ while keeping the operational constraints fixed ($\xi^{ru} = 0.3$ and $\xi^{as} = 0.3$). Figure~\ref{figEC-Comp2} shows the Pareto frontiers for three levels of FDA performance ($k \in \{L,M,H\}$), filtering for policies that reject fewer than 15\% of safe devices to reflect more feasible cases in terms of implementation. As shown in Figure~\ref{figEC-Comp2}, our policy's frontiers are strictly dominant over the ML-only policy across all levels of $k$. The performance gap is largest in the desirable operational region, where the rejection rate of safe devices is relatively low. We observe that the performance of our policy improves with higher levels of $k$. In other words, when the FDA committee becomes more effective at rejecting deferred unsafe devices, our policy’s partial deferral strategy becomes increasingly beneficial. In practical terms, the ML-only policy is most effective when its risk estimations are highly accurate, where the ML-only policy can make accurate decisions for hard cases. However, in many real-world scenarios, including ours, risk estimation models perform reasonably well but lack uniform accuracy across all cases. In such settings, selectively deferring difficult cases to human experts can significantly enhance decision-making and lead to a better-performing policy overall.

\begin{figure} [tbh] 
    \centering
    \includegraphics[width=0.6\linewidth]{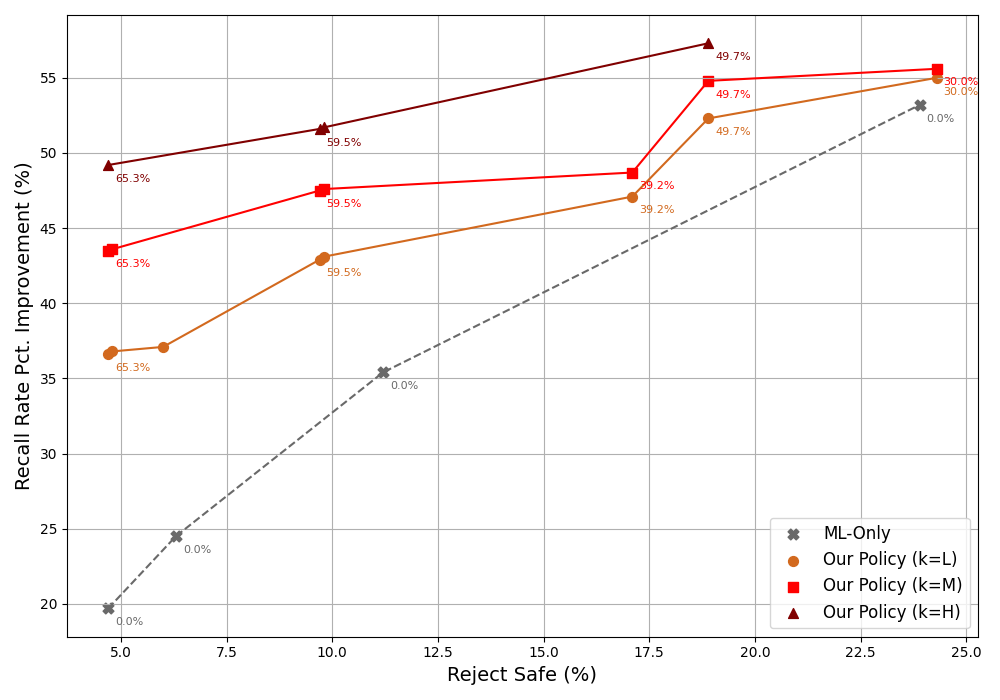}
    \caption{Comparison of the performance of our policy vs the ML-only policy in the testing set under ($\lambda \in [0.1,0.9]$, $\xi^{ru} = 0.3$, and $\xi^{as} = 0.3$). The annotated text on each curve indicates the corresponding FDA's workload.} 
    \label{figEC-Comp2}
\end{figure}

\color{black}
\section{Estimation of Replacement Costs of Recalled Medical Devices}{ \label{crosswalk}

In this section, we discuss how the replacement costs of recalled medical devices are calculated. As noted by the OIG report, the replacement cost of recalled medical devices cannot be tracked to individual devices based solely on claims data, as Medicare claim forms do not contain a field for reporting medical device-specific information (\citealt{USdephealth2017}). Consequently, our estimate of replacement costs is specific to the medical specialty, not individual devices. We determined the medical specialty for over 99\% of the 1,351 unique HCPCS codes/descriptions in the MDMEDS 2013-2020 data by first identifying keywords in the device classification names available in the 510(k) submission data for each medical specialty and then matching those keywords to the HCPCS descriptions in the MDMEDS data.
For example, an HCPCS description containing the word “ostomy,” which is a procedure used to treat various diseases of the urinary or digestive systems, was classified as having the medical specialty Gastroenterology/Urology. In a slightly more intricate case, HCPCS descriptions containing the words “glucose monitor” or “glucose” and “monitor” were classified as having the medical specialty Clinical Chemistry. Table~\ref{table10} shows the crosswalk between keywords and medical specialties.

\begin{table}[hbt]
  \centering \footnotesize
  \caption{Medical Specialties and Keywords}
  \begin{tabular}{|l|p{10cm}|}
    \hline
    \textbf{Medical Specialty} & \textbf{Keyword(s)} \\
    \hline
    Anesthesiology &
    ``aerosol'' and ``compressor'',
    ``airway'',
    ``breathing circuits'',
    ``cough'',
    ``face mask'',
    ``nasal cannula'',
    ``nasal mask'',
    ``nebulization'',
    ``nebulizer'',
    ``oropharyngeal'',
    ``oxygen'',
    ``positive expiratory pressure'',
    ``respiratory'',
    ``tracheal suction'',
    ``ventilator'' \\
    \hline
    Cardiovascular &
    ``defibrillator'',
    ``pneumatic compression device'' \\
    \hline
    Clinical Chemistry &
    ``calibrator solution'',
    ``glucose monitor'',
    ``glucose'' and ``monitor'' \\
    \hline
    Dental &
    ``osteogenesis'' \\
    \hline
    Gastroenterology/Urology &
    ``bladder'',
    ``cervical'',
    ``drainage bag'',
    ``indwelling catheter'',
    ``insertion tray'' and ``catheter'',
    ``leg strap'',
    ``male'' and ``catheter'',
    ``ostomy'',
    ``parenteral'',
    ``pelvic floor'',
    ``stoma cap'',
    ``urethral'',
    ``urinary'' \\
    \hline
    General \& Plastic Surgery &
    ``adhesive'',
    ``bandage'',
    ``chest wall'',
    ``collagen'' and ``wound'',
    ``compression'' and ``wrap'',
    ``dressing'',
    ``gauze'',
    ``lancet'',
    ``skin barrier'',
    ``sterile water'',
    ``tape'',
    ``tubing'' and ``pump'',
    ``ultraviolet'' and ``therapy'',
    ``wound'' \\
    \hline
    General Hospital &
    ``ambulatory infusion pump'',
    ``bath'',
    ``bed'',
    ``canister'' and ``pump'',
    ``chair'',
    ``compression stocking'',
    ``compression'' and ``garment'',
    ``compressor'' and ``for equipment'',
    ``drug infusion'',
    ``footplate'',
    ``footrests'',
    ``heel loop'',
    ``infusion pump'',
    ``insulin'',
    ``irrigation'',
    ``iv pole'',
    ``lubricant'',
    ``mattress'',
    ``transfer device'',
    ``urinal'' and ``jug-type'' \\
    \hline
    Neurology &
    ``conductive garment'',
    ``nerve stimulation'' \\
    \hline
    Physical Medicine &
    ``armrest'',
    ``cane'',
    ``commode chair'',
    ``crutches'',
    ``electrical'' and ``stimulator'',
    ``flexion'',
    ``foot'' and ``density insert'',
    ``foot'' and ``shoe molded'',
    ``heat pad'',
    ``inlay'' and ``shoe'',
    ``knee'' and ``exercise'',
    ``leg'' and ``compressor'',
    ``neuromuscular stimulator'',
    ``patient lift'',
    ``patient support system'',
    ``patient transfer'',
    ``pneumatic'' and ``compressor'',
    ``rear wheel'',
    ``traction'' and ``cervical'',
    ``trapeze'',
    ``vehicle'',
    ``walker'',
    ``wheelchair'' \\
    \hline
  \end{tabular} \label{table10}
\end{table}

Once the medical specialties for the HCPCS codes were established, we calculated the average Medicare allowed amount per medical specialty. This calculation was based on the total supplier claims, which reflects the number of products ordered by the referring provider. The Medicare allowed amount includes the amount Medicare paid, the deductible and coinsurance amounts owed by the beneficiary, as well as any amount owed by a third-party payer (\citealt{CMS1}). We were able to calculate the average Medicare allowed amount for over 75\% of the devices in our test dataset based on their respective medical specialties. However, for the devices for which we were unable to compute the average Medicare allowed amount, we faced a challenge in assigning a specific medical specialty to them. Among them, Radiology devices account for 13.7\% of all devices in the test dataset and the remaining specialties cumulatively account for less than 8.5\% of devices. We believe that the chief reason we could not calculate Medicare costs for these specialties is that the underlying 510(k) devices are not single-use devices, expended on a single patient. In the case of radiology, we examined the majority of 510(k) cleared products between 2013 and 2020 and determined that these devices were either imaging software or multi-use equipment used in providing radiology services such as radiosurgery or radiotherapy. For the purposes of our impact assessment, we created low and high estimates of the average Medicare allowed amount when we could not directly calculate the average Medicare allowed amount. The low estimate is the lowest average Medicare allowed amount across all of the specialties. The high estimate is the weighted average Medicare allowed amount across all specialties, with weights derived from the frequency of each medical specialty in our testing dataset. 

\color{black}
\section{Proofs}{
All proofs for lemmas, propositions, and theorems are given below.
\begin{repeatproposition}[Proposition 1.] \label{aux-prop1}
For any $\theta \in (0,1)$ and a pair of $h(\xi^{ru})$ and $\ell(\xi^{as})$ in the Auxiliary Problem, we have:
\begin{enumerate} [label=(\alph*)]
\renewcommand\labelenumi{\normalfont(\alph{enumi})}
    \item if $h(\xi^{ru}) \;>\; \ell(\xi^{as})$, then $(\ell(\xi^{as}), h(\xi^{ru}))$ is the unique optimal solution,
    \item if $h(\xi^{ru}) \;<\; \ell(\xi^{as})$, then the problem is infeasible, 
    \item if $h(\xi^{ru}) \;=\; \ell(\xi^{as})$, then $\ell \;=\; h \;=\; h(\xi^{ru}) \;=\; \ell(\xi^{as})$ is the single threshold optimal solution.
\end{enumerate}
\end{repeatproposition}

\proof{Proof of Proposition \ref{aux-prop1}:}

We prove each case separately.

\noindent \textbf{Case (a).} We first find an optimal solution, and then we prove its uniqueness. When $h(\xi^{ru}) \;>\; \ell(\xi^{as})$, the polyhedral feasible region of the problem has three extreme points: $(\ell(\xi^{as}), \ell(\xi^{as}))$, $(\ell(\xi^{as}), h(\xi^{ru}))$, and $(h(\xi^{ru}), h(\xi^{ru}))$. The objective values corresponding to these extreme points are $\ell(\xi^{as})(1-2\theta)$, $-\theta \, \ell(\xi^{as}) + (1-\theta) h(\xi^{ru})$, and $h(\xi^{ru})(1-2\theta)$, respectively. For any $\theta \in (0,1)$, we have  $-\theta \, \ell(\xi^{as}) + (1-\theta) h(\xi^{ru}) \; > \; \ell(\xi^{as})(1-2\theta)$ because $h(\xi^{ru}) \;>\; \ell(\xi^{as})$ and $\theta < 1$. Also, for any $\theta \in (0,1)$, we have $-\theta \, \ell(\xi^{as}) + (1-\theta) h(\xi^{ru})  \; > \; h(\xi^{ru})(1-2\theta)$ because $h(\xi^{ru}) \;>\; \ell(\xi^{as})$ and $\theta > 0$. Accordingly, $(\ell(\xi^{as}), h(\xi^{ru}))$ is an optimal solution for the problem. 

Next, we use a contradiction argument to prove that $(\ell(\xi^{as}), h(\xi^{ru}))$ is the unique optimal solution. Suppose that there is another optimal solution $(\bar{\ell},\bar{h}) \ne (\ell(\xi^{as}), h(\xi^{ru}))$. According to the polyhedral feasible region, there are two possible scenarios: (1) $\bar{h} \; < \; h(\xi^{ru})$ and $\bar{\ell} \; \ge \; \ell(\xi^{as})$, or (2) $\bar{h} \; \le \; h(\xi^{ru})$ and $\bar{\ell} \; > \; \ell(\xi^{as})$. In both scenarios, we have:
$$  -\theta \; \bar{\ell} + (1-\theta) \; \bar{h}  \; < \;  -\theta \; \ell(\xi^{as}) + (1-\theta) \; h(\xi^{ru}). $$
This contradicts the optimality of $(\bar{\ell},\bar{h})$. Thus, we conclude that $(\ell(\xi^{as}), h(\xi^{ru}))$ is the unique optimal solution. 

\noindent \textbf{Case (b).} The condition of $h(\xi^{ru}) \; < \; \ell(\xi^{as})$, results in $h \; < \; \ell$ which contradicts the requirement of $0 \; \le \; \ell \; \le \; h \; \le \; 1$. Thus, the problem is infeasible.

\noindent \textbf{Case (c).} This is a direct result of Case (a).

\QED
\endproof

\begin{repeattheorem}[Theorem 1.] \label{aux-thm1}
For any $\theta \in (0,1)$ and $\lambda \in (0,1)$, and a pair of $h(\xi^{ru})$ and $\ell(\xi^{as})$, we have:
\begin{enumerate} [label=(\alph*)]
\renewcommand\labelenumi{\normalfont(\alph{enumi})}
    \item if $h(\xi^{ru}) \;>\; \ell(\xi^{as})$, then the two-threshold optimal solution of the Auxiliary Problem is optimal in the Relaxed Problem,
    \item if $h(\xi^{ru}) \;<\; \ell(\xi^{as})$, then both problems are infeasible, 
    \item if $h(\xi^{ru}) \;=\; \ell(\xi^{as})$, then the single threshold optimal solution of the Auxiliary Problem is optimal in the Relaxed Problem.
\end{enumerate}
\end{repeattheorem}

\proof{Proof of Theorem \ref{aux-thm1}:}
First, we highlight that both problems have the same polyhedral feasible region. Note that $h,\ell \in [0,1]$ and interval $[0,1]$ is convex and compact. Supremum is attained since $\mathbb{P}(\delta^{+} \ge h)$ is continuous and weakly decreasing in $h$. Similarly, infimum is attained since $\mathbb{P}(\delta^{-} \le \ell)$ is continuous and weakly increasing in $\ell$. Accordingly, we have:
\begin{align*}
\mathbb{P}(\delta^{+} \ge h) \; \ge \; \xi^{ru} \;\;  \Longleftrightarrow \;\; h \; \le \; h(\xi^{ru}), 
\; \; \text{and} \quad \mathbb{P}(\delta^{-} \le \ell) \; \ge \; \xi^{as} \;\; \Longleftrightarrow \;\; \ell \; \ge \; \ell(\xi^{as}).
\end{align*}

\color{black}
Hence, any feasible solution of the Auxiliary Problem is also a feasible solution to the Relaxed Problem. 

Next, we show that both problems have the same optimal solutions in each case. 

\noindent \textbf{Case (a).} By Proposition \ref{prop1}, when $h(\xi^{ru}) \;>\; \ell(\xi^{as})$, we have that $(\ell(\xi^{as}), h(\xi^{ru}))$ is the unique optimal solution of the Auxiliary Problem for any $\theta \in (0,1)$. The objective function of the Relaxed Problem is the convex combination of $\mathbb{P}(\delta^{+} \le \ell)$ and $\mathbb{P}(\delta^{-} \ge h)$, which are monotone functions. Since $\mathbb{P}(\delta^{+} \le \ell)$ is weakly increasing in $\ell$, we have $\mathbb{P}(\delta^{+} \le \ell) \; \ge \; \mathbb{P}(\delta^{+} \le \ell(\xi^{as}))$ for any $\ell \; \ge \; \ell(\xi^{as})$. Similarly, since $\mathbb{P}(\delta^{-} \ge h)$ is weakly decreasing in $h$, we have $\mathbb{P}(\delta^{-} \ge h) \; \ge \; \mathbb{P}(\delta^{-} \ge h(\xi^{ru}))$ for any $ h \; \le \; h(\xi^{ru})$. Accordingly, for any feasible pair of $\ell$ and $h$, we have: 
$$  \lambda \; \mathbb{P}(\delta^{+} \le \ell) + (1-\lambda) \; \mathbb{P}(\delta^{-} \ge  h) \; \ge \; \lambda \; \mathbb{P}(\delta^{+} \le \ell(\xi^{as})) + (1-\lambda) \; \mathbb{P}(\delta^{-} \ge h(\xi^{ru})).$$
Therefore, we can conclude that $(\ell(\xi^{as}), h(\xi^{ru}))$ is also the optimal solution of the Relaxed Problem.

\color{black}
\noindent \textbf{Case (b).} By Proposition \ref{prop1}, when $h(\xi^{ru}) \;<\; \ell(\xi^{as})$, the Auxiliary Problem is infeasible. The Relaxed Problem is also infeasible since both problems are restricted to the same set of constraints. 

\noindent \textbf{Case (c).} This is a direct result of Case (a).

\QED
\endproof

\begin{repeattheorem}[Theorem 2.] \label{aux-thm2}
For the \textcolor{black}{Primary Problem}, we have: 
\begin{enumerate} [label=(\alph*)]
\renewcommand\labelenumi{\normalfont(\alph{enumi})}
    \item If $h(\xi^{ru}) \;>\; \ell(\xi^{as})$ and  the workload constraint \eqref{c:workloadCDF} holds for $\ell \,=\,  \ell(\xi^{as})$ and $h \,=\, h(\xi^{ru})$, then $(\ell(\xi^{as}), h(\xi^{ru}))$ is an optimal solution,
    \item  If $h(\xi^{ru}) \;>\; \ell(\xi^{as})$ and the workload constraint \eqref{c:workloadCDF} does not hold for $\ell \,=\,  \ell(\xi^{as})$ and $h \,=\, h(\xi^{ru})$, then for any given $\epsilon > 0$, our nested search algorithm returns an $\epsilon$-optimal solution after at most 
    $$ \mathcal{O} \left( \frac{ \bar{L} \left(  h(\xi^{ru}) - \ell(\xi^{as}) \right)}{\epsilon}  \; \log_2 \left( \frac{ 2 \bar{L}_{\phi} (1-\lambda) \left(h(\xi^{ru})-\ell(\xi^{as}) \right)}{\epsilon}  \right) \right)$$ 
    iterations, where $\lambda \in (0,1)$, $\bar{L}_{\phi}$ and $\bar{L}_{h}$ are, respectively, upper bounds on the common Lipschitz constant of $\phi^{\pm}(\cdot)$ and the Lipschitz constant of $h^{*}(\ell)$, and  $\bar{L} = \bar{L}_{\phi} \left(\lambda + (1-\lambda)\bar{L}_h \right)$. Setting $\Delta \; = \; \epsilon/[2\bar{L}]$ and $\zeta \;=\; \epsilon/[2(1-\lambda)\bar{L}_{\phi}] $ guarantees $\epsilon$-optimality and ensures that the workload constraint is satisfied within a tolerance of $\epsilon/[2 (1-\lambda)]$,
    \item If $h(\xi^{ru}) \;<\; \ell(\xi^{as})$, then the problem is infeasible,   
    \item If $h(\xi^{ru}) \;=\; \ell(\xi^{as})$, then the problem has a single threshold solution $\ell^* \;=\; h^* \;=\; h(\xi^{ru}) \;=\; \ell(\xi^{as}).$
\end{enumerate}
\end{repeattheorem}
\color{black}

\proof{Proof of Theorem  \ref{aux-thm2}:} 
We prove each case separately. 

\noindent \textbf{Case (a).} \textcolor{black}{Since $(\ell(\xi^{as}), h(\xi^{ru}))$ meets the FDA's workload constraint \eqref{c:workloadCDF}, the \textcolor{black}{Primary Problem} becomes equivalent to the Relaxed Problem.} By Theorem \ref{thm1}, if $h(\xi^{ru}) \;>\; \ell(\xi^{as})$, then the optimal solution of the Auxiliary Problem is optimal in the Relaxed Problem. 
Consequently, $(\ell(\xi^{as}), h(\xi^{ru}))$ is also an optimal solution of the \textcolor{black}{Primary Problem}.

\noindent \textbf{Case (b).} In this case, 
$(\ell(\xi^{as}), h(\xi^{ru}))$ is not a feasible solution because it violates 
the FDA's workload constraint by assumption. In particular, the problem has the 
following feasible region:
\begin{align*}
\Theta = \Big\{ (\ell,h) \; \text{s.t.} \;
&q\left(\phi^+(h)-\phi^+(\ell)\right)
+(1-q)\left(\phi^-(h)-\phi^-(\ell)\right)\le \rho, 
 h\le h(\xi^{ru}), 
\ell\ge \ell(\xi^{as})
\Big\}.
\end{align*}

We show that our nested search finds a near-optimal solution within a finite 
number of steps. To do so, we bound the sub-optimality gap
$J(\ell^{NS},h^{NS})-J(\ell^*,h^*)$, where $J(\cdot)$ is the objective 
function of the \textcolor{black}{Primary Problem}.

Recall that
\[
g(\ell,h)
=
q\bigl(\phi^+(h)-\phi^+(\ell)\bigr)
+
(1-q)\bigl(\phi^-(h)-\phi^-(\ell)\bigr)
-\rho.
\]
Under the conditions of Case (b), there exists an optimal solution at which the workload constraint binds, 
so that $g(\ell^*,h^*)=0.$

Let $\ell_k = \max\{\ell\in\mathcal{L}:\ell\le \ell^*\}$,
that is, $\ell_k$ is the grid point immediately to the left of $\ell^*$. 
Since the grid has step size $\Delta$, we have $
0\le \ell^*-\ell_k \le \Delta.$

We next show that $\ell_k\in\mathcal{L}_{\mathrm{feas}}$. Since $g(\ell,h)$ 
is nonincreasing in $\ell$ and nondecreasing in $h$, and since
$\ell_k\le \ell^*$ and $h^*\le h(\xi^{ru})$, we have
\begin{align*}
g(\ell_k,h(\xi^{ru})) \ge g(\ell^*,h(\xi^{ru})) \ge g(\ell^*,h^*) =0.
\end{align*}
Moreover, we have $ g(\ell_k,\ell_k)=-\rho\le 0$. 
Therefore, there exists $h^*(\ell_k)\in[\ell_k,h(\xi^{ru})]$ such that
$g(\ell_k,h^*(\ell_k))=0$. Hence, $\ell_k\in\mathcal{L}_{\mathrm{feas}}$.

By design of our algorithm, $\ell^{NS}$ is selected such that it yields a 
value less than or equal to the value at any other feasible grid point. 
Therefore, $
J(\ell^{NS},h^{NS})
\le
J(\ell_k,h^*_{\zeta}(\ell_k)).$

Accordingly, we decompose the optimality gap as follows:
\begin{align*}
J(\ell^{NS},h^{NS})-J(\ell^*,h^*)
&\le
J(\ell_k,h^*_{\zeta}(\ell_k))-J(\ell^*,h^*) \\
&=
\underbrace{
\left[
J(\ell_k,h^*_{\zeta}(\ell_k))
-
J(\ell_k,h^*(\ell_k))
\right]
}_{\text{Sub-optimality due to Bisection Search}} 
 +
\underbrace{
\left[
J(\ell_k,h^*(\ell_k))
-
J(\ell^*,h^*)
\right]
}_{\text{Sub-optimality due to Grid Search}},
\end{align*}
where $h^*(\ell)$ is the exact root of equation
\eqref{eq:nested_workload} for a feasible grid point
$\ell\in\mathcal{L}_{\mathrm{feas}}$.

We proceed with the proof in two main steps.

\noindent \textbf{Step 1 (Sub-optimality due to Bisection Search for $h$).}
The first term can be expanded as:
\begin{align*}
J(\ell_k,h^*_{\zeta}(\ell_k))
-
J(\ell_k,h^*(\ell_k))
=
(1-\lambda)
\left(
\phi^-(h^*(\ell_k))
-
\phi^-(h^*_{\zeta}(\ell_k))
\right).
\end{align*}
Assuming $\phi^-(\cdot)$ is $L_{\phi}$-Lipschitz, the above term can be 
upper bounded as follows:
\[
J(\ell_k,h^*_{\zeta}(\ell_k))
-
J(\ell_k,h^*(\ell_k))
\le
(1-\lambda)L_{\phi}
\left|
h^*(\ell_k)-h^*_{\zeta}(\ell_k)
\right|.
\]

Let $h^{(n)}(\ell)$ denote the solution of the bisection after $n$ iterations 
for $\ell\in\mathcal{L}_{\mathrm{feas}}$. The bisection search generates a 
sequence of $\{h^{(n)}(\ell)\}_{n=1}^{\infty}$ approximating the root of the 
workload constraint such that (\citealt{gautschi2011numerical}):
\[
\left|
h^{(n)}(\ell)-h^*(\ell)
\right|
\le
\frac{h(\xi^{ru})-\ell(\xi^{as})}{2^n},
\qquad
\forall n\ge 1.
\]
Selecting the tolerance as $\zeta =
\frac{h(\xi^{ru})-\ell(\xi^{as})}{2^n}$,
the number of iterations to reach this tolerance is at most
$ \log_2
\left(
\frac{h(\xi^{ru})-\ell(\xi^{as})}{\zeta}
\right)$.
Thus, the first term of the sub-optimality gap is at most
$(1-\lambda)L_{\phi}\zeta$ after $
\log_2
\left(
\frac{h(\xi^{ru})-\ell(\xi^{as})}{\zeta}
\right)$ 
iterations.

Next, we show that the solution of the bisection search, $h^{NS}$, satisfies 
the workload constraint within a tolerance level. Recall that for a given 
$\ell$, we have
\[
g(\ell,h)
=
q\bigl(\phi^+(h)-\phi^+(\ell)\bigr)
+
(1-q)\bigl(\phi^-(h)-\phi^-(\ell)\bigr)
-\rho.
\]
Also, by definition, $
g(\ell^{NS},h^*(\ell^{NS}))=0$. Therefore,
\begin{align*}
|g(\ell^{NS},h^{NS})|
&=
\left|
g(\ell^{NS},h^{NS})
-
g(\ell^{NS},h^*(\ell^{NS}))
\right| \\
&=
\Big|
q\bigl(
\phi^+(h^{NS})
-
\phi^+(h^*(\ell^{NS}))
\bigr) 
+
(1-q)\bigl(
\phi^-(h^{NS})
-
\phi^-(h^*(\ell^{NS}))
\bigr)
\Big| \\
&\le
q
\left|
\phi^+(h^{NS})
-
\phi^+(h^*(\ell^{NS}))
\right| 
+
(1-q)
\left|
\phi^-(h^{NS})
-
\phi^-(h^*(\ell^{NS}))
\right| \\
&\le
qL_{\phi}
\left|
h^{NS}-h^*(\ell^{NS})
\right|
+
(1-q)L_{\phi}
\left|
h^{NS}-h^*(\ell^{NS})
\right| \\
&\le
qL_{\phi}\zeta
+
(1-q)L_{\phi}\zeta \\
&=
L_{\phi}\zeta.
\end{align*}

\noindent \textbf{Step 2 (Sub-optimality due to Grid Search for $\ell$).}
Let $
F^*(\ell)
=
\lambda\phi^+(\ell)
+
(1-\lambda)
\left(
1-\phi^-(h^*(\ell))
\right)$. By this definition, we have:
\[
J(\ell_k,h^*(\ell_k))
-
J(\ell^*,h^*)
=
F^*(\ell_k)-F^*(\ell^*).
\]

We now establish the Lipschitz constant for $F^*(\ell)$. We assume 
$\phi^+(\cdot)$ is $L_{\phi}$-Lipschitz. Then the first term,
$\lambda\phi^+(\ell)$, is $(\lambda L_{\phi})$-Lipschitz by the multiplication 
property. For the second term, by assumption, $\phi^-(\cdot)$ is 
$L_{\phi}$-Lipschitz and $h^*(\ell)$ is $L_h$-Lipschitz. By the composition 
property, the composite function $
1-\phi^-(h^*(\ell))$ is $(L_{\phi}L_h)$-Lipschitz. It follows from the scalar multiplication 
property that the entire second term is
$((1-\lambda)L_{\phi}L_h)$-Lipschitz. By the sum property, the Lipschitz 
constant for $F^*(\ell)$ is the sum of the constants of its terms: $
\lambda L_{\phi}
+
(1-\lambda)L_{\phi}L_h
=
L.$

Accordingly, we can upper bound the sub-optimality due to the grid search as
follows:
\[
J(\ell_k,h^*(\ell_k))
-
J(\ell^*,h^*)
\le
L|\ell_k-\ell^*|
\le 
L\Delta,
\]
where the inequality holds by the Lipschitz property of $F^*(\cdot)$ and the 
fact that $ 0\le \ell^*-\ell_k \le \Delta$.

As a result, our nested search yields an $\epsilon$-optimal solution
$(\ell^{NS},h^{NS})$ after at most
$ \mathcal{O}
\left(
\frac{h(\xi^{ru})-\ell(\xi^{as})}{\Delta}
\log_2
\left(
\frac{h(\xi^{ru})-\ell(\xi^{as})}{\zeta}
\right)
\right)$ 
iterations. The workload feasibility is satisfied within
$L_{\phi}\zeta$, and the sub-optimality is bounded by
$ \epsilon
=
L\Delta
+
(1-\lambda)L_{\phi}\zeta$. 

Next, we compute the maximum number of iterations required for any given
$\epsilon>0$. While $L_{\phi}$ and $L_h$ are generally unknown, we can use 
known upper bounds,
$\bar L_{\phi}\ge L_{\phi}$ and $\bar L_h\ge L_h$. 
Therefore, for a given value of $\epsilon$, we can run our nested search 
algorithm with
\[
\Delta
=
\frac{\epsilon}{2\bar L},
\qquad
\text{and}
\qquad
\zeta
=
\frac{\epsilon}
{2(1-\lambda)\bar L_{\phi}}.
\]

Under these settings, the sub-optimality is:
\begin{align*}
J(\ell^{NS},h^{NS})-J(\ell^*,h^*)
&\le
L\Delta
+
(1-\lambda)L_{\phi}\zeta =
\frac{L}{\bar L}\frac{\epsilon}{2}
+
\frac{L_{\phi}}{\bar L_{\phi}}\frac{\epsilon}{2} \le \epsilon.
\end{align*}

Similarly, the workload feasibility is satisfied with the desired tolerance:
\[
|g(\ell^{NS},h^{NS})|
\le
L_{\phi}\zeta
=
\frac{L_{\phi}}{\bar L_{\phi}}
\frac{\epsilon}{2(1-\lambda)}
\le
\frac{\epsilon}{2(1-\lambda)}.
\]

Accordingly, our nested search returns an $\epsilon$-optimal solution while 
keeping the workload feasibility within
$\epsilon/[2(1-\lambda)]$. Substituting the values of $\Delta$ and $\zeta$, 
the maximum number of iterations required is:
\[
\mathcal{O}
\left(
\frac{
\bar L\left(h(\xi^{ru})-\ell(\xi^{as})\right)
}{
\epsilon
}
\log_2
\left(
\frac{
2(1-\lambda)\bar L_{\phi}
\left(h(\xi^{ru})-\ell(\xi^{as})\right)
}{
\epsilon
}
\right)
\right).
\]

\color{black}
\noindent \textbf{Case (c).} The condition of $h(\xi^{ru}) \; < \; \ell(\xi^{as})$ results in $h \; < \; \ell$, which contradicts the requirement of $0 \; \le \; \ell \; \le \; h \; \le \; 1$ in the \textcolor{black}{Primary Problem}. Thus, the problem is infeasible.

\noindent \textbf{Case (d).} This can be shown by following a closely analogous argument to Case (a).

\QED
\endproof

}
}

\section{Nested Search Algorithm} \label{EC:nested search}
This section presents the pseudocode of Algorithm \ref{alg:nested_search} for the nested search procedure discussed in Section~\ref{nested search}. This algorithm is employed to determine the risk thresholds $(\ell, h)$ when the workload constraint is binding.

\setcounter{algorithm}{0} 
\begin{algorithm}[tbh]
\caption{Nested Search Algorithm}
\label{alg:nested_search}
\begin{algorithmic}[1]

\State \textbf{Input:} Empirical CDFs $\phi^+$ and $\phi^-$; parameters $q$, $\lambda$, $\rho$, $\xi^{as}$, $\xi^{ru}$; grid step $\Delta$; tolerance $\zeta$
\State \textbf{Initialize:} Set of feasible candidates $\mathcal{L}_{\text{feas}} \gets \emptyset$
\State Construct a grid $\mathcal{L}$ over $[\ell(\xi^{as}), h(\xi^{ru}))$ using the grid step $\Delta$
\For{each $\ell \in \mathcal{L}$}
    \State Set $h_L \gets \ell$, $h_R \gets h(\xi^{ru})$
    \State Compute $g(\ell, h_L)$ and $g(\ell, h_R)$, where $g(\ell, h) = q(\phi^+(h) - \phi^+(\ell)) + (1-q)(\phi^-(h) - \phi^-(\ell)) - \rho$

    \If{$g(\ell, h_L) > 0$ or $g(\ell, h_R) < 0$}
        \State \textbf{continue} to next $\ell$ 
    \EndIf
    \While{$|h_R - h_L| > \zeta$}
        \State $h_{\text{mid}} \gets (h_L + h_R)/2$
        \State Compute $g(\ell, h_{\text{mid}})$ 
        \If{$g(\ell, h_{\text{mid}}) < 0$}
            \State $h_L \gets h_{\text{mid}}$
        \Else
            \State $h_R \gets h_{\text{mid}}$
        \EndIf
    \EndWhile
    \State Set $h^{*}_{\zeta}(\ell) \gets (h_L + h_R)/2$
    \State Add $\ell$ to the set of feasible candidates $\mathcal{L}_{\text{feas}}$
    \State Compute $F(\ell) = \lambda  \phi^+(\ell) + (1 - \lambda)(1 - \phi^-(h^{*}_{\zeta}(\ell)))$ and record it
\EndFor
\State Select the thresholds as:
$$ \ell^{NS} = \arg\min_{\ell \in \mathcal{L}_{\text{feas}}} F\big( \ell \big),
\quad
h^{NS} = h^{*}_{\zeta}\big( \ell^{NS} \big) $$
\end{algorithmic}
\end{algorithm}

\end{APPENDICES}

\end{document}